\documentclass[runningheads]{llncs}

\usepackage{eccv}
\usepackage{eccvabbrv}
\usepackage{graphicx}
\usepackage{booktabs}

\usepackage{amsmath}
\usepackage{amssymb}
\usepackage{mathtools}

\usepackage{placeins}
\usepackage{booktabs}
\usepackage{multirow}
\usepackage{xcolor} 
\usepackage{bm}

\usepackage{algorithm}
\usepackage{algpseudocode}

\definecolor{mycolor_green}{HTML}{E6F8E0}
\definecolor{mycolor_white}{HTML}{F7F7F7}

\newcommand{\myparagraph}[1]{\noindent\textbf{#1}\hspace{0.4em}}

\usepackage[accsupp]{axessibility}
\usepackage{hyperref}
\usepackage{orcidlink}

\begin{document}

\setlength{\textfloatsep}{9pt} 
\setlength{\floatsep}{9pt} 
\setlength{\intextsep}{9pt}

\title{AnchorWeave: World-Consistent Video Generation with Retrieved Local \\ Spatial Memories} 

\titlerunning{AnchorWeave}

\author{Zun Wang\inst{1}\orcidlink{0009-0005-9502-050X} \and
Han Lin\inst{1}\orcidlink{0000-0002-1303-8636} \and
Jaehong Yoon\inst{3}\orcidlink{0000-0002-9653-9590} \and
Jaemin Cho\inst{2,4}\orcidlink{0000-0002-1558-6169} \and
Yue Zhang\inst{1}\orcidlink{0000-0003-2153-6536} \and
Mohit Bansal\inst{1}\orcidlink{0009-0009-2965-5354}}

\authorrunning{Z.~Wang et al.}

\institute{University of North Carolina at Chapel Hill, USA \and
Johns Hopkins University, USA \and
Nanyang Technological University, Singapore \and
Allen Institute for AI, USA\\
\url{https://zunwang1.github.io/AnchorWeave}}

\maketitle

\begin{abstract}

Maintaining spatial world consistency over long horizons remains a central challenge for camera-controllable video generation.
Existing memory-based approaches often condition generation on globally reconstructed 3D scenes by rendering anchor videos from the reconstructed geometry in the history.
However, reconstructing a global 3D scene from multiple views inevitably introduces cross-view misalignment, as pose and depth estimation errors cause the same surfaces to be reconstructed at slightly different 3D locations across views.
When fused, these inconsistencies accumulate into noisy geometry that contaminates the conditioning signals and degrades generation quality.
We introduce \textbf{AnchorWeave}, a memory-augmented video generation framework that replaces a single misaligned global memory with multiple clean local geometric memories and learns to reconcile their cross-view inconsistencies. 
To this end, AnchorWeave performs coverage-driven local memory retrieval aligned with the target trajectory and integrates the selected local memories through a multi-anchor weaving controller during generation.
Extensive experiments demonstrate that AnchorWeave significantly improves long-term scene consistency while maintaining strong visual quality, with ablation and analysis studies further validating the effectiveness of local geometric conditioning, multi-anchor control, and coverage-driven retrieval. 
We provide more qualitative demonstration videos in the supplementary material via an anonymized local webpage.
\end{abstract}

\section{Introduction}

Building video world models that can consistently generate and revisit the same environments over long horizons under user control (such as keyboard directions~\cite{bruce2024genie, parkerholder2024genie2, genie3}, camera viewpoints~\cite{wang2024motionctrl, he2024cameractrl, chen2025learning}, or human/robot actions~\cite{guo2025ctrl}
remains a central challenge in generative modeling. Recent image-to-video diffusion and Transformer-based models~\cite{wan2025, kong2024hunyuanvideo, hacohen2026ltx} have significantly improved short-term coherence and controllability, enabling clip-by-clip autoregressive generation~\cite{huang2025selfforcing, zhang2025frame, chendiffusion} that follows camera trajectories. However, such models usually lack persistent memory mechanisms, causing them to lose consistency with previously generated content when revisiting earlier regions of the environment~\cite{yu2025context}.

\begin{figure*}[t]
    \centering
    \includegraphics[width=0.95\linewidth]{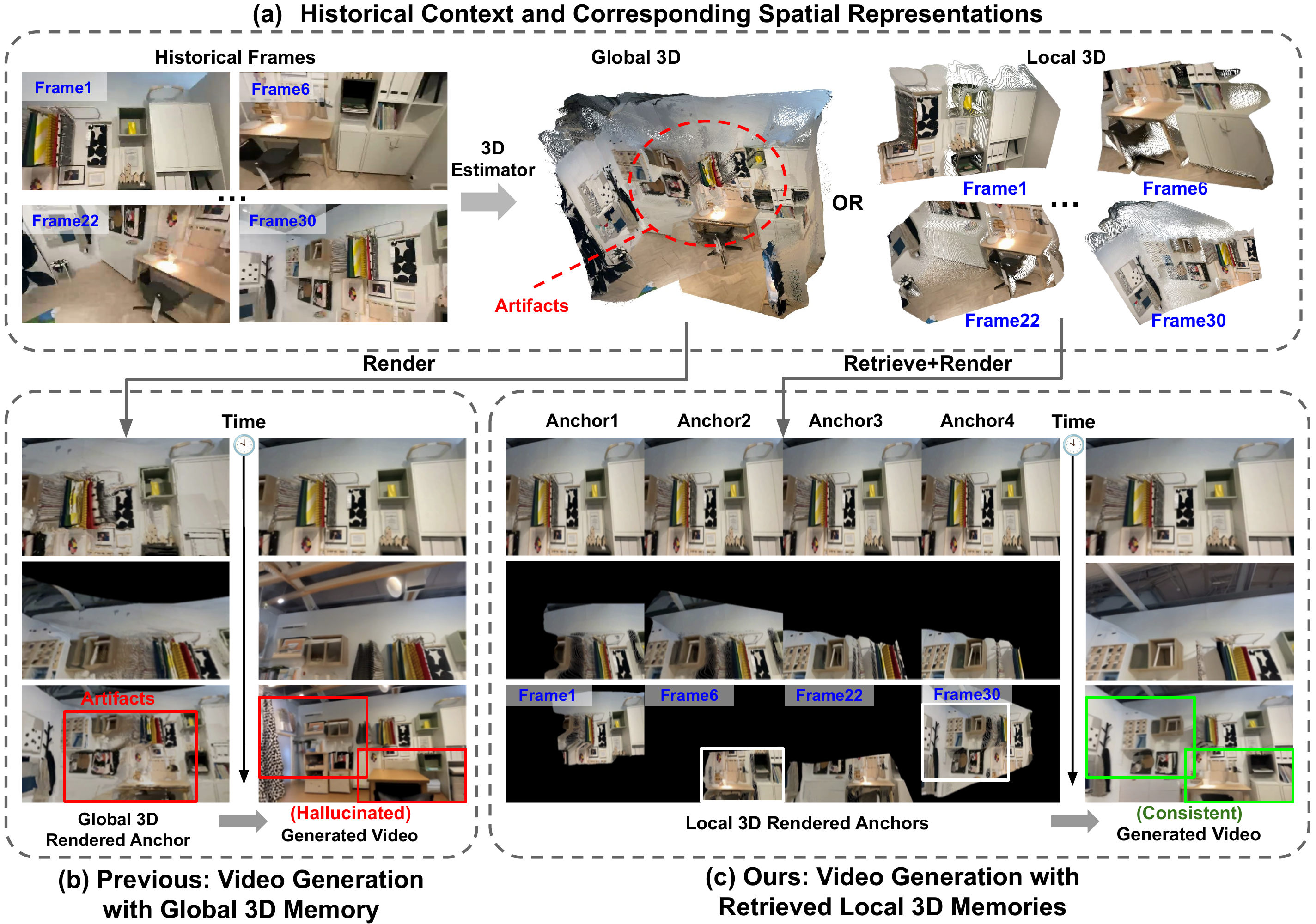}
    \caption{
    Global 3D reconstruction accumulates cross-view misalignment, introducing artifacts in the reconstructed geometry ((a), middle), which propagate to the generated video as hallucinations ((b), red boxes).
In contrast, per-frame local geometry inherently avoids cross-view misalignment and therefore remains clean ((a), right).
Conditioning on multiple retrieved local geometric anchors, AnchorWeave maintains strong spatial consistency with the historical frames ((c), white and green boxes).
}
\label{fig:teaser}
\end{figure*}

To improve long-horizon world-consistent video generation, recent works~\cite{zhao2025spatia, li2025magicworld, wu2025spmem} introduce explicit memory mechanisms that reconstruct a global 3D scene from historical video clips. 
Specifically, they estimate camera poses and per-frame geometry (e.g., depth maps), and fuse them into a unified 3D representation such as a global point cloud. Generation is then conditioned on anchor videos (i.e., videos rendered from the reconstructed 3D scene under target camera trajectories) to enable spatial consistency.
However, even state-of-the-art 3D reconstruction models struggle to maintain accurate global geometry under dense, long-horizon views, making precise cross-view alignment difficult. 
Small pose or depth estimation errors cause the same surfaces to be reconstructed at slightly different 3D locations across views (\Cref{fig:teaser}(a) middle artifacts); when fused, these inconsistencies introduce structured noise into rendered anchor videos, leading to ghosting or drift artifacts that degrade generation quality (e.g., inconsistent or hallucinated content in \Cref{fig:teaser}(b)).

We propose \textbf{AnchorWeave}, a 3D-informed, memory-aware video generation framework that avoids relying on a globally fused scene representation. 
Our key insight is that cross-view misalignment artifacts can be mitigated by replacing a single global 3D memory with per-frame local point cloud memories, as illustrated in \Cref{fig:teaser}(c). 
Since local point clouds do not accumulate ghosting or drift from multi-view fusion, they provide cleaner geometric signals for rendering-based conditioning. 
While these rendered anchors may still exhibit residual misalignment, we treat their reconciliation during generation as a learnable problem and resolve it through a learned multi-anchor weaving module.
Specifically, given a sequence of previously generated or observed frames, AnchorWeave estimates and maintains per-frame local geometry as spatial memory for long-horizon generation. 
To synthesize new content under a target camera trajectory, AnchorWeave introduces a coverage-driven memory retrieval formulation that iteratively selects the local memory that maximizes additional visibility coverage along the target trajectory.
The selected memories are rendered as multiple anchor videos for conditioning. 
A multi-anchor weaving controller with shared cross-anchor attention and pose-guided fusion then \emph{weaves} these anchors into a unified control signal, enabling effective multi-anchor conditioning while mitigating residual geometric inconsistencies.
The newly generated frames are updated to the history for subsequent generation, enabling AnchorWeave to iteratively extend the video through the update–retrieve–generate loop.

We compare AnchorWeave with multiple baselines spanning different memory conditioning paradigms, including global point cloud and implicit cross-view conditioning, on RealEstate10K and DL3DV under a history-conditioned video generation setting.
AnchorWeave significantly improves visual quality and long-horizon scene consistency, while generalizing well to open-domain images and scenes.
Ablation studies further validate the contribution of each component, including local geometric memory, coverage-driven retrieval, and the geometry-conditioned multi-anchor controller, supporting the effectiveness of conditioning on local geometry and learning to handle misalignment.

\section{Related Work}

\myparagraph{Camera-controllable video generation.}
Recent advances in video generation with camera controllability span text-to-video (T2V), image-to-video (I2V), and video-to-video (V2V) settings. 
Early approaches primarily inject explicit camera parameterizations (e.g., Pl\"ucker embeddings or pose tokens) into video diffusion models for conditioning~\cite{wang2024motionctrl, hou2024learning, bahmani2024vd3d, bahmani2024ac3d, sun2024dimensionx, he2025cameractrl, zheng2024cami2v, xu2024camco, watson2024controlling, yuegosim, li2025realcam, hecameractrl, zhou2025stable, li2024nvcomposer,liang2025wonderland,liao2025thinking,li2025recamdriving,wang2025taming,zhang2025unified}. 
To improve geometric faithfulness, subsequent works introduce structured 3D guidance, such as conditioning on rendered point clouds or anchor videos~\cite{yu2024viewcrafter,popov2025camctrl3d,hou2024training, ren2025gen3c, zheng2025vidcraft3, seo2024genwarp, cao2025uni3c, muller2024multidiff, liu2024reconx, zhang2024recapture, zhang2025i2v3d, zhou2024latent, yang2025omnicam, bernal2025precisecam,huang2025voyager,lee20253d,zhang2025world}. 
Another line of research formulates camera control through trajectory tracking or motion encoding as intermediate guidance~\cite{jin2025flovd, feng2024i2vcontrol, xiao2024trajectory, gu2025diffusion,burgert2025go,luo2025camclonemaster,wu2025geometry,fortier2025gimbaldiffusion}. 
In the V2V regime, some methods rely on test-time optimization or per-scene fine-tuning~\cite{you2024nvs, song2025worldforge, zhang2024recapture, park2025zero4d}, while others construct large-scale paired data from simulators such as Unreal Engine5, Kubric, or Animated Objaverse~\cite{bai2025recammaster, bai2024syncammaster, greff2022kubric, van2024generative, deitke2023objaverse, wu2024cat4d, gao2024cat3d, yu20244real, wang20244real,huang2025spacetimepilot,van2026anyview}. 
Recent Approaches also incorporate structured 3D priors for controllable V2V generation~\cite{bian2025gs, yu2025trajectorycrafter,hu2025ex,yang2026neoverse,lu2025see4d,xie2026lavr,chen2025postcam,pan2025diff4splat}. 
Despite substantial progress, existing camera-controllable video models are typically memoryless: they condition on per-frame parameters, rendered anchors, or short local context, but lack an explicit, robust memory mechanism to maintain long-horizon spatial consistency under complex camera motions. 
In contrast, we focus on equipping camera control models with a geometry-aware and retrieval-based memory design, enabling robust long-horizon consistency and controllability beyond single-shot or short-context conditioning.

\myparagraph{Memory-augmented video generation.}
Recent video world models extend long-horizon consistency by incorporating memory beyond the limited context window of current generators. 
One line of work adopts retrieval-based mechanisms, treating past frames as external memory and retrieving relevant history for future synthesis~\cite{xiao2025video4spatial,chou2025captain,yu2025context,li2025vmem,xiao2025worldmem,gu2025long,guo2024infinitydrive}. 
Another line preserves spatial structure through explicit 3D memory constructed from historical inputs, such as maintaining global point clouds to support revisit consistency and camera control~\cite{li2025wonderplay,yu2025wonderworld,li2025magicworld,zhou2025learning,liu2025dynamem,zhao2025spatia,wu2025spmem,wu2025geometry,li2025magicworld}. 
A third direction leverages recurrent or state-space models to enhance long-horizon temporal modeling~\cite{po2025long,savov2025statespacediffuser,zhang2025test,dalal2025one,lee2025enhancing,hong2024slowfast}. 
Several works also improve consistency or long-term quality by redesigning attention or latent conditioning in autoregressive or diffusion-based models~\cite{he2025matrix,chen2024diffusion,song2025history,wu2025pack,wang2026worldcompass,xiang2025pan,team2026advancing, samuel2026fast,chen2026past,ma2026flow,huang2025self,yang2025longlive,zhang2025frame,mao2025yume,ye2025yan,liu2025rolling,sun2025worldplay,RelicWorldModel2025}.
Among them, the closest are 3D-based approaches including SPMem~\cite{wu2025spmem}, Spatia~\cite{zhao2025spatia}, MagicWorld~\cite{li2025magicworld} that rely on global 3D point cloud as memory. Instead of noisy global reconstruction, we represent memory as local geometric anchors and perform multi-anchor joint conditioning for more stable long-horizon consistency.

\section{Methodology: AnchorWeave}
\textbf{AnchorWeave} is a world-consistent video generation framework that leverages retrieved local spatial memories for long-horizon, camera-controlled generation.
It builds upon a standard video diffusion backbone (\Cref{preliminary}) augmented with explicit spatial memory.
Instead of fusing observations into a single global 3D memory (\textit{i.e.,} global point cloud),
AnchorWeave maintains a collection of per-frame local geometric memories, each represented as a local point cloud with an associated camera pose (\Cref{Local Geometric Memory Construction}).
Given this memory bank, relevant memories are selected for a target camera trajectory using a coverage-driven, 3D-aware retrieval strategy (\Cref{sec:memory_retrieval}) and rendered as anchor videos.
These anchor videos are integrated into the diffusion process through a Multi-anchor Weaving Controller, which jointly reasons over multiple anchors via shared attention and pose-guided fusion (\Cref{Multi-anchor Weaving Controller}).
Finally, following \cite{wu2025spmem} that condition on global memory, AnchorWeave operates in an iterative update--retrieve--generate loop, where newly generated frames are converted into local geometric memories to support long-horizon and consistent generation (\Cref{Long-horizon Generation Loop with Iterative Memory Update}).

\subsection{Preliminary: Video Diffusion Models}
\label{preliminary}
We use standard DiT-based latent diffusion models (LDMs)~\cite{rombach2022high} as backbone.
Specifically, given an RGB video with $F$ frames,
$\bm{x}\in\mathbb{R}^{F\times 3 \times H \times W}$,
a pretrained video autoencoder first compresses the input into a lower-dimensional latent tensor:
$\bm{z}=\mathcal{E}(\bm{x})\in\mathbb{R}^{F\times C' \times H' \times W'}$,
where $H'$ and $W'$ denote the spatially downsampled resolution.
To enable generative modeling, noise is injected into the latent representation through a predefined scheduling strategy, such as DDPM~\cite{ho2020denoising} or Flow Matching~\cite{lipmanflow}, producing a noisy latent $\bm{z}_t$ at timestep $t$.
A conditional diffusion network
$\bm{\mathcal{F}_\theta}(\bm{z}_t, t, \bm{c}_{\text{text/img}})$
is then trained to recover the underlying clean signal by reversing this corruption process,
where the model is conditioned on the diffusion timestep and auxiliary inputs, including textual descriptions $\bm{c}_{\text{text}}$ for text-to-video generation, and extra camera intrinsics and anchor videos $\bm{c}_{\text{cam}}$ for video generation with camera control.
Training is performed by minimizing the following objective:
$
\mathcal{L}_{\text{LDM}} =
\mathbb{E}_{\bm{z},\,\bm{\epsilon}\sim\mathcal{N}(0,\bm{I}),\,t}
\left\|
\bm{\epsilon} -
\bm{\epsilon}_{\theta}(\bm{z}_t, t, \bm{c}_{\text{text+cam}})
\right\|_2^2$
where $\bm{\epsilon}$ represents the sampled Gaussian perturbation and
$\bm{\epsilon}_{\theta}$ denotes the noise estimate produced by the model.
This objective is used to train all diffusion components.

\begin{figure*}[t]
    \centering
    \includegraphics[width=\linewidth]{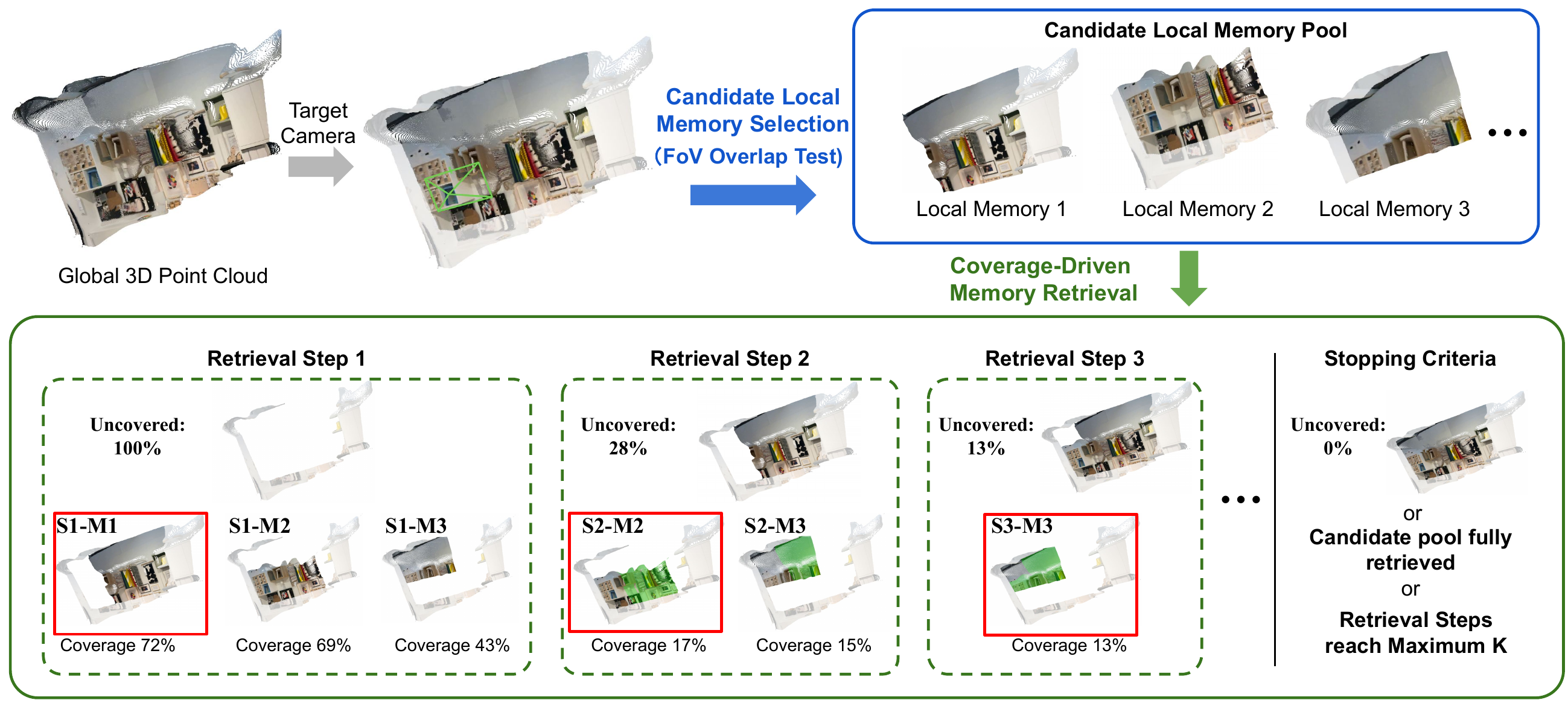}
    \caption{
    Coverage-driven memory retrieval pipeline.
Given a target camera, we first select local memories whose camera FoVs partially overlap with the target camera view to form a candidate memory pool. At each retrieval step, we greedily select the memory that maximizes the newly covered visible area.
Points invisible to the target camera are highlighted in \colorbox{mycolor_white}{white}
(For example, in the top row, the white regions in the second point cloud (compared to the first) correspond to areas outside the target camera’s field of view as invisible regions).
S$i$–M$j$ denotes memory $j$ selected at retrieval step $i$; the red box indicates the retrieved memory. 
In S$i$-M$j$, regions already covered by previously retrieved memories are marked in \colorbox{mycolor_green}{green}. Only newly covered regions retain their original RGB colors. No green regions appear in S1-M$j$ since 1st-step's coverage is empty.
Retrieval terminates when uncovered region is 0\%, the retrieval budget $K$ is exhausted, or the remaining memory pool is empty. For clarity, coverage is computed with a single frame here, while in practice is aggregated over multiple frames per chunk.   
}
\label{fig:retrieval_pipeline}
\end{figure*}

\subsection{Local Geometric Memory Construction}
\label{Local Geometric Memory Construction}
We represent \textbf{spatial memory as a set of per-frame local point clouds} rather than a single global 3D model, thereby avoiding the accumulation and propagation of cross-view misalignment artifacts.
The construction proceeds as follows.
We assume access to a history context consisting of previously observed video frames, which can come from real-world inputs or earlier generated segments. These frames serve as the visual context from which spatial memory is constructed and updated throughout long-horizon generation.
Using this history context, we estimate per-frame local geometry and camera pose with a pretrained 3D reconstruction model (e.g., TTT3R~\cite{chen2025ttt3r}). Each frame's memory is represented as a local point cloud together with its camera pose, which is transformed into a shared world coordinate system and stored as an independent spatial memory entry. This update can be performed incrementally only for newly generated frames in practice.

\subsection{Coverage-driven Memory Retrieval}
\label{sec:memory_retrieval}
With a large pool of history local memories, we aim to retrieve a compact subset that can effectively guide generation for a target camera trajectory.
We illustrate the retrieval pipeline in~\Cref{fig:retrieval_pipeline} and provide the complete retrieval procedure in the Appendix.

\myparagraph{Retrieval pipeline.}
Given a target camera trajectory of length $T$ for the next video segment, we divide it into $T/D$ temporal chunks, each spanning $D$ 
frames,
and perform memory retrieval independently for each chunk.
We formulate memory retrieval as a \textbf{visibility coverage maximization} problem: the selected local memories should jointly maximize coverage along the target camera trajectory within each chunk.
Specifically, as shown in the upper part of~\Cref{fig:retrieval_pipeline}, for each chunk, we first construct a candidate local memory pool using a coarse field-of-view (FoV) overlap test to filter out clearly irrelevant local memories.
Among the remaining candidates, we iteratively select the local memory that provides the largest new coverage of the currently uncovered region along the chunk trajectory, shown in the bottom part of~\Cref{fig:retrieval_pipeline},
Retrieval stops when the visible region is fully covered, the candidate pool is exhausted, or the number of retrieved memories reaches the maximum $K$.
This greedy strategy yields a compact and complementary set of local memories per chunk, avoiding redundant memories that offer little additional guidance.

\myparagraph{Anchor-video construction and pose computation.}
For each chunk, we retrieve up to $K$ local memories.
For each retrieved memory, we render anchor views along the target camera trajectory within the chunk, producing anchor clips of length $D$ frames.
If fewer than $K$ memories are retrieved, we pad the remaining anchor slots with fully-invisible anchor clips to keep a fixed number of $K$ anchors per chunk.

For the rendered anchors, we compute the retrieved-to-target relative camera pose sequence, defined as the rigid transformations between the camera pose at which the local memory was captured and the target camera poses within the chunk.
Anchor clips corresponding to the same retrieval index are then concatenated across chunks, resulting in $K$ anchor videos of length $T$ with temporally aligned retrieved-to-target pose trajectories.
These anchor videos and their pose sequences are used as conditioning signals for generation.

\begin{figure*}[t]
    \centering
    \includegraphics[width=\linewidth]{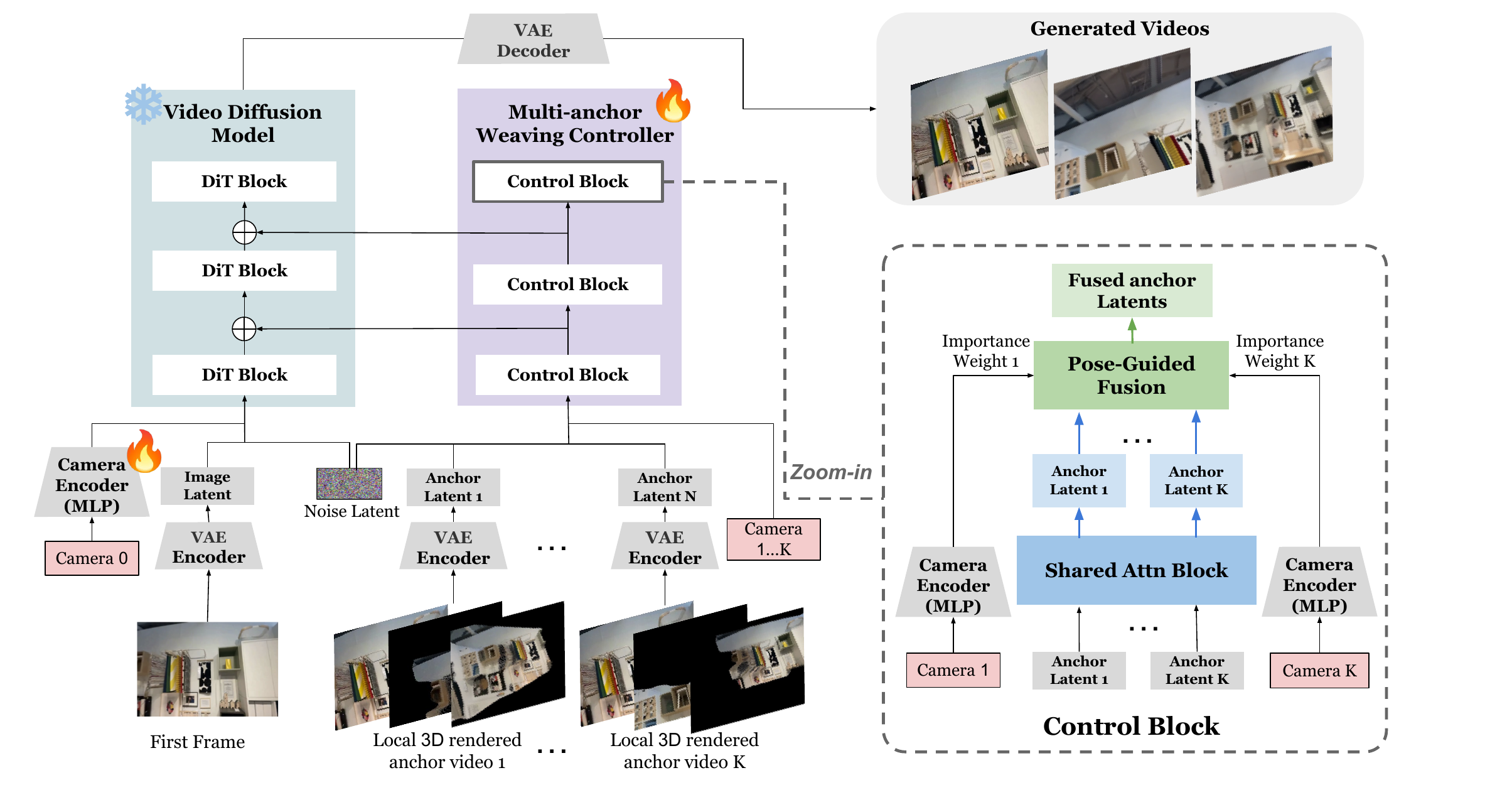}
    \caption{Architecture of multi-anchor weaving controller.  Anchors are encoded and jointly processed by a shared attention block, followed by camera-pose-guided fusion to produce a unified control signal injected into the backbone model. Camera 1 to $K$ represent the retrieved-to-target camera poses for the 1 to $K$ anchor videos, where each denotes the relative pose between the camera associated with a retrieved local point cloud and the target camera, measuring their viewpoint proximity. Camera 0 is the relative target camera trajectory.
    }
    \label{fig:controller}
\end{figure*}

\subsection{Generation with Multi-Anchor Weaving Controller}
\label{Multi-anchor Weaving Controller}
Given $K$ retrieved anchor videos, we condition generation on their latents together with retrieved-to-target relative camera poses, and the target camera pose, enabling the model to \emph{weave} multiple anchor conditions into a coherent and spatially consistent output video.

\myparagraph{Overall architecture.}
As shown in \Cref{fig:controller}, each retrieved anchor video is first encoded into latent features using the same 3D VAE as the backbone diffusion model. These anchor latents are then fed into a multi-anchor controller, composed of a stack of DiT-based ControlNet blocks that operate alongside the backbone model.
At each denoising layer, we enable joint multi-anchor modeling through two key designs: (1) a shared multi-anchor attention module that jointly processes all anchor features, and (2) a camera-pose-guided fusion module that aggregates anchor features into a single control signal. The resulting fused control feature is injected into the corresponding layer of the video diffusion backbone to provide geometric guidance. Details of these designs are presented below.

\myparagraph{Shared multi-anchor attention.}
Instead of processing anchors independently, we jointly process the $K$ anchor latents using a shared-attention block (\Cref{fig:controller} inside the control block).
Concretely, we reshape the $K$ anchor latent sequences with shape $L_a \times C_a$  into a single sequence of shape $1 \times (K \cdot L_a) \times C_a$ ($L_a$ denotes the number of latent tokens per anchor and $C_a$ is the hidden dimension), and apply joint attention over the concatenated tokens.
This joint attention enables information exchange across anchors, allowing the controller to aggregate complementary geometric evidence while suppressing contradictory or noisy signals, rather than committing to any single anchor rendering.

\myparagraph{Pose-aware importance estimation and fusion.}
Since not all anchors contribute equally to a target view, we condition anchor fusion on the retrieved-to-target relative camera poses (\Cref{fig:controller} right, camera 1 to $K$).
At each control layer, these camera poses are encoded by a learnable MLP to produce per-anchor importance weights, which reflect the geometric proximity between the target view and the cameras associated with the retrieved local point clouds used to render the anchors.
Using these importance weights, we perform a weighted sum over the anchor-conditioned features to get a single fused control feature, which is injected into the corresponding backbone block to provide geometric guidance. 

\myparagraph{Target camera pose control.}
We further condition the model on the target camera trajectory (denoted as camera 0 in \Cref{fig:controller}) to handle cases where geometric memories provide limited guidance, such as early generation stages with rapid camera motion that moves the target view beyond all anchor visibility.
The target camera pose is encoded by a trainable camera encoder to produce camera embeddings,
which are injected into the backbone DiT blocks as explicit camera-motion control signals.
This complementary signal improves camera following under long-horizon generation.

Overall, the multi-anchor weaving controller explicitly conditions generation on geometry through joint attention, pose-conditioned fusion, and target camera pose control, transforming imperfect anchors into a coherent conditioning signal.

\subsection{Long-horizon Generation with Updated Memories}
\label{Long-horizon Generation Loop with Iterative Memory Update}
Each iteration of the components described in Sections \ref{Local Geometric Memory Construction}-\ref{Multi-anchor Weaving Controller} produces one video segment. Repeating this process enables long-horizon video generation via an \emph{update–retrieve–generate} loop in AnchorWeave.
\textbf{(1) Update (\Cref{Local Geometric Memory Construction}):} Starting from an initial frame or short segment, we estimate per-frame local geometry and initialize the spatial memory bank.
\textbf{(2) Retrieve (\Cref{sec:memory_retrieval}):} Given a target camera trajectory, we retrieve relevant local memories using our coverage-driven memory retrieval and condition the model on both the target pose and retrieved anchors.
\textbf{(3) Generate (\Cref{Multi-anchor Weaving Controller}):} The backbone model generates the next video segment under the control of the multi-anchor geometry-aware controller.
The newly generated frames are then added to the memory bank, enabling memory-augmented generation over  long trajectories.

\section{Experiments}

\subsection{Experiment Setup}

\myparagraph{Implementation details.}
\label{implementation}
We train AnchorWeave on 10K videos sampled from RealEstate10K~\cite{zhou2018stereo} and DL3DV~\cite{ling2024dl3dv}.
For each video, per-frame local geometry is estimated using TTT3R~\cite{chen2025ttt3r} and stored as local point-cloud memories.
During training, we use chunk length $D=8$, retrieving memories every 8 frames with at most $K=4$ point clouds for conditioning.
To improve robustness, we randomly sample from the candidate memory pool as condition augmentation.
For memory retrieval, the first rendered frame is always selected as the initial anchor, following the standard anchor-video convention~\cite{zhang2024recapture,yu2025trajectorycrafter,ren2025gen3c}.
We apply AnchorWeave to two DiT-based backbones: CogVideoX-I2V-5B~\cite{yang2024cogvideox} and Wan2.2-TI2V-5B~\cite{wan2025}.
The multi-anchor controller is injected into the first third of backbone layers and applied during the first 90\% of denoising steps.
We use random frame masking on anchor videos to improve robustness to imperfect geometric guidance.
The model is trained with Adam (lr=$2\times10^{-4}$) for 10K steps with batch size 8.
Additional details are provided in the Appendix.

\myparagraph{Evaluation setup and dataset.}
We evaluate the model's ability to condition on historical context under a \emph{partial-revisit} setting. We sample videos with large camera motion (with substantial viewpoint changes between frames) from RealEstate10K~\cite{zhou2018stereo} and DL3DV~\cite{ling2024dl3dv}, collecting a total of 500 videos for evaluation.
For each video, we sample 70 frames, uniformly select 49 frames as the target segment, and use the remaining 21 frames as historical context available for retrieval.
This enables direct quantitative and qualitative comparison against ground-truth (GT) frames. 
In addition, we include long-horizon qualitative examples using keyboard-style camera actions starting from a single open-domain initial frame, assessing 
long-term controllability and scene consistency. 

\myparagraph{Evaluation metrics.}
Following~\cite{wu2025spmem}, we use PSNR and SSIM to measure reconstruction fidelity, computed on predicted and GT frames.
For perceptual quality, the VBench protocol~\cite{huang2024vbench} is adopted, including Subject/Background Consistency, Motion Smoothness, Temporal Flickering, Aesthetic /Imaging Quality.

\begin{table*}[t]
\centering
\caption{Quantitative results on RealEstate10K and DL3DV under partial-revisit evaluation.
The best numbers are in \textbf{bold} and the second best numbers are \underline{underlined}. The total quality score is the average of all quality dimensions. $\dagger$ denotes our reimplementation based on CogVideoX with the same data as ours (due to not open-sourced). $K$ denotes the maximum retrieval per chunk.}
\label{tab:main_re10k_mira}
\small
\setlength{\tabcolsep}{4pt}
\resizebox{1.0\linewidth}{!}{
\begin{tabular}{lccccccc cc}
\toprule
\multirow{3}{*}{\textbf{Method}} &
\multicolumn{7}{c}{\textbf{Quality Score ($\uparrow$)}} &
\multicolumn{2}{c}{\textbf{Consistency Score ($\uparrow$)}} \\
\cmidrule(lr){2-8} \cmidrule(lr){9-10}
& \multirow{2}{*}{Total} &
\multicolumn{1}{c}{Subject} &
\multicolumn{1}{c}{Bg} &
\multicolumn{1}{c}{Motion} &
\multicolumn{1}{c}{Temporal} &
\multicolumn{1}{c}{Aesthetic} &
\multicolumn{1}{c}{Imaging} &
\multirow{2}{*}{PSNR} &
\multirow{2}{*}{SSIM} \\
& &
Consist &
Consist &
Smooth &
Flicker &
Quality &
Quality & & \\
\midrule

ViewCrafter~\cite{yu2024viewcrafter} & 
78.58 & 85.50 & 90.11 & 90.91 & 91.36 & 46.71 & \textbf{66.92} & 16.17 & 0.5240 \\

Gen3C~\cite{ren2025gen3c} & 
77.11 & 84.22 & 89.08 & 93.74 & 90.43 & 45.25 & 59.92 & 18.23 & 0.5926 \\

TrajCrafter~\cite{yu2025trajectorycrafter} & 
76.34 & 83.56 & 88.94 & 91.43 & 88.06 & 45.85 & 60.19 & 16.96 & 0.5767 \\

EPiC~\cite{wang2025epic} & 
77.77 & 85.22 & 90.08 & 94.74 & 90.43 & 45.25 & 60.92 & 17.42 & 0.5704 \\

Context-as-Memory$^\dagger$~\cite{yu2025context} & 
78.07 & 85.08 & 89.94 & 92.36 & 90.21 & 46.97 & 63.88 & 17.91 & 0.5884 \\

SPMem$^\dagger$~\cite{wu2025spmem} & 
76.85 & 84.52 & 89.31 & 91.84 & 89.67 & 45.61 & 60.12 & 17.25 & 0.5710 \\

SEVA~\cite{zhou2025stable} & 
79.66 & 87.64 & 92.31 & 94.31 & 91.00 & 48.90 & 63.78 & \textbf{21.13} & 0.6711 \\

\midrule
AnchorWeave (CogVideoX-5B, $K{=}1$) & 
80.07 & 87.70 & 92.51 & 96.33 & 93.05 & 47.88 & 62.94 & 19.01 & 0.6145 \\

AnchorWeave (CogVideoX-5B) & 
80.30 & 87.89 & 92.34 & 96.21 & \underline{93.90} & 47.87 & 63.61 & 20.96 & \underline{0.6727} \\

AnchorWeave (Wan2.2-5B, $K{=}1$) & 
\underline{80.76} & \underline{88.05} & \textbf{92.78} & \textbf{96.58} & 93.32 & \textbf{49.40} & 64.45 & 19.60 & 0.6180 \\

AnchorWeave (Wan2.2-5B) & 
\textbf{80.98} & \textbf{88.25} & \underline{92.60} & \underline{96.45} & \textbf{94.15} & \underline{49.35} & \underline{65.10} & \underline{21.04} & \textbf{0.6739} \\

\bottomrule
\end{tabular}
}
\end{table*}

\subsection{Comparison with Prior Methods}
\label{com prior}

\myparagraph{Baselines.}
We compare AnchorWeave with state-of-the-art camera-controllable, memory-augmented video generation methods,
 including Gen3C~\cite{ren2025gen3c}, SEVA~\cite{zhou2025stable}, ViewCrafter~\cite{yu2024viewcrafter}, 
 TrajCrafter~\cite{yu2025trajectorycrafter}, EPiC~\cite{wang2025epic}, Context-as-Memory~\cite{yu2025context}, and SPMem~\cite{wu2025spmem}.
Gen3C, TrajCrafter, ViewCrafter, and EPiC are designed to condition generation on a single anchor video.
To provide strong baselines, we construct the anchor video for these methods using our retrieval strategy with 
$K$=1, where the retrieved local memory corresponds to the local point cloud that maximizes spatial coverage for each chunk. This allows these baselines to condition on the most informative retrieved local memory, in the same spirit as our method.
We additionally include a single-anchor variant of AnchorWeave ($K=1$, one retrieval per chunk) to directly compare with these baselines under identical conditioning settings.
Context-as-Memory and SEVA directly take multi-view history as conditions for memory consistency, while SPMem conditions generation on global point cloud renderings together with selected key historical frames.
Note that, as Context-as-Memory and SPMem are not open-sourced, we reimplement them on the CogVideoX backbone using the same training data as our approach for fair comparison. Additional implementation details are provided in the Appendix.

\myparagraph{Quantitative comparison.}
\Cref{tab:main_re10k_mira} shows that AnchorWeave achieves the best overall performance across both reconstruction and perceptual quality metrics.
While AnchorWeave and SEVA~\cite{zhou2025stable} obtain comparable consistency scores, AnchorWeave exhibits significantly better temporal quality, including smoother motion and reduced flickering, benefiting from preserving the pretrained video backbone’s motion prior.
Compared to single-anchor baselines, our single-anchor variant ($K=1$) consistently achieves higher quality and consistency, indicating that our design extracts more fine-grained geometric details from anchor conditioning and corrects residual errors.
Our method also consistently surpasses memory-based baselines, including Context-as-Memory and SPMem.
Moreover, the full AnchorWeave further improves over its $K=1$ variant, demonstrating the effectiveness of multiple retrieval to improve consistency. Notably, AnchorWeave already outperforms all baselines when built on the CogVideoX backbone, and replacing CogVideoX with Wan2.2 leads to further performance gains. We also evaluate camera controllability compared to baselines in the Appendix.

\begin{figure*}[!t]
    \centering
        \includegraphics[width=\linewidth]{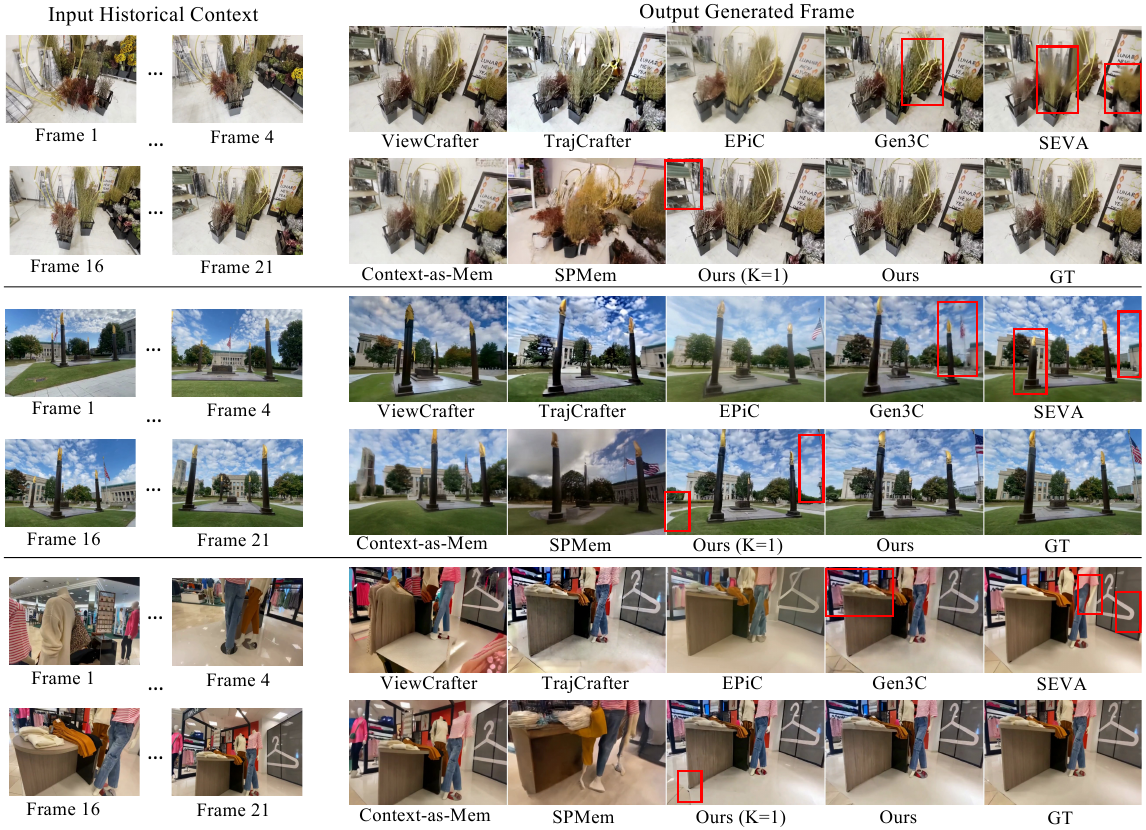}
    \caption{Qualitative comparison with baselines on DL3DV. K=1 means one retrieval per chunk. Baseline methods suffer from spatial drift and inconsistency in details. In contrast, AnchorWeave~(ours) maintains consistency for multiple cases. GT and the historical context are shown for reference. For clarity, representative misaligned regions are highlighted with red boxes for strong baselines (e.g., SEVA).
    }
    \label{fig:qual_com}
\end{figure*}

\myparagraph{Qualitative comparison.}
\Cref{fig:qual_com} shows qualitative comparisons under the partial-revisit setting. We present one representative frame for each case. Additional qualitative comparisons under free-form, keyboard-controlled camera trajectories with historical context are provided in the Appendix.
Overall,
AnchorWeave maintains more consistent scene structure when revisiting, while baseline methods exhibit visible artifacts or misaligned details.
Gen3C, EPiC, and TrajCrafter often produce frames with degraded visual quality, including noticeable blurring or structural distortions.
ViewCrafter is prone to hallucinated content (2nd/3rd example), while SEVA~\cite{zhou2025stable} occasionally fails to preserve fine-grained details (see highlighted red-box regions). 
Context-as-Memory maintains content and scene consistency but lacks precise camera control, often resulting in noticeable deviations from the GT viewpoint.
SPMem conditions on rendered global point clouds, which lead to severe hallucinations that remain difficult to mitigate even with additional key-frame conditioning.
In contrast, AnchorWeave preserves structural details more reliably through dense, clean local memory conditioning.
The single-anchor variant ($K=1$) sometimes exhibits misalignment due to limited retrieval (see red boxes), whereas the full model achieves stronger consistency by weaving multiple anchors.
Overall, AnchorWeave produces results that are visually closer to the ground truth.

\begin{table}[t]
\centering
\caption{Results of conditioning on global or local Memory.
}
\label{tab:ablation_geometry_controller}
\small
\setlength{\tabcolsep}{6pt}
\resizebox{0.6\linewidth}{!}{
\begin{tabular}{lcc}
\toprule
\textbf{Method} & \textbf{PSNR} $\uparrow$ & \textbf{SSIM} $\uparrow$ \\
\midrule
AnchorWeave w/ global point cloud & 16.31 & 53.45 \\
\midrule
AnchorWeave w/ local point clouds (ours) & \textbf{20.96} & \textbf{67.27} \\
\bottomrule
\end{tabular}
}
\end{table}

\begin{table}[t]
\centering
\caption{Effect of number of retrievals. We vary the number of retrievals $K$ per chunk, resulting in at most $6K$ retrieved frames (i.e., $6$, $12$, and $24$ for $K$=$1,2,4$, respectively). Retrieved frames are not necessarily unique across chunks and duplicates may occur.}
\label{tab:ablation_num_retrievals}
\small
\setlength{\tabcolsep}{6pt}
\resizebox{0.55\linewidth}{!}{
\begin{tabular}{ccc}
\toprule
\textbf{ Max. \# retrieved frames} & \textbf{PSNR} $\uparrow$ & \textbf{SSIM} $\uparrow$ \\
\midrule
6  & 19.01 & 0.6145 \\
12 &  20.01 & 0.6435 \\ 
24 & 20.96 & 0.6727 \\
\bottomrule
\end{tabular}
}
\end{table}

\begin{figure*}[!t]
    \centering
    \includegraphics[width=\linewidth]{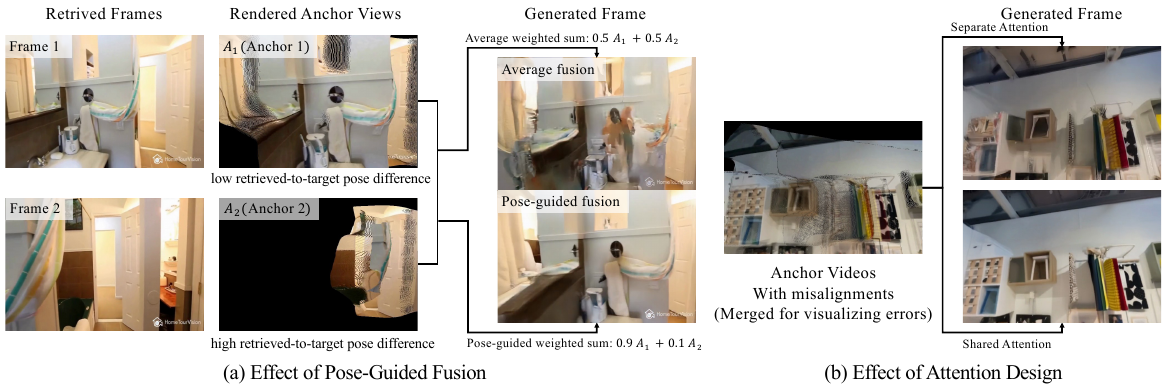}
    \caption{Ablation study.
(a) Pose-guided fusion suppresses misaligned anchors and reduces artifacts compared to simple averaging.
(b) Joint attention outperforms separate attention, enabling coherent multi-anchor aggregation.}
    \label{fig:ablation}
\end{figure*}

\subsection{Ablations and Analyses}

We conduct a series of ablations to analyze the design choices of AnchorWeave, including the use of global vs.\ local 3D memory, pose-conditioned anchor fusion, joint vs.\ separate attention in the controller, and the number of retrieved frames.
The Appendix further complements these analyses with targeted diagnostics: it visualizes how global point-cloud fusion accumulates drift, stress-tests repeated revisit consistency over 2400 frames, and discusses dynamic-content handling and representative failure modes.

\myparagraph{Effect of global vs. local 3D memory.}
We compare global and local 3D memories as conditioning signals for training the controller.
In the global-memory setting, we render a single anchor from the fused global 3D memory, while in the local-memory setting, multiple anchors are rendered from local memories as conditions.
The results in \Cref{tab:ablation_geometry_controller} show that local 3D memory yields improvements in both PSNR and SSIM, confirming the advantage of local geometry for memory conditioning.

\myparagraph{Effect of pose-conditioned anchor fusion.} 
As shown in ~\Cref{fig:ablation}(a), simply averaging multiple retrieved anchor views leads to visible artifacts in the generated video when the anchors exhibit different pose deviations from the target view.
In contrast, pose-conditioned fusion assigns higher weights to anchors with smaller pose discrepancies, effectively suppressing misaligned anchors and improving visual quality, demonstrating that explicitly accounting for retrieved-to-target pose differences is crucial for robust multi-anchor conditioning.

\begin{figure*}[!t]
    \centering
    \includegraphics[width=0.99\linewidth]{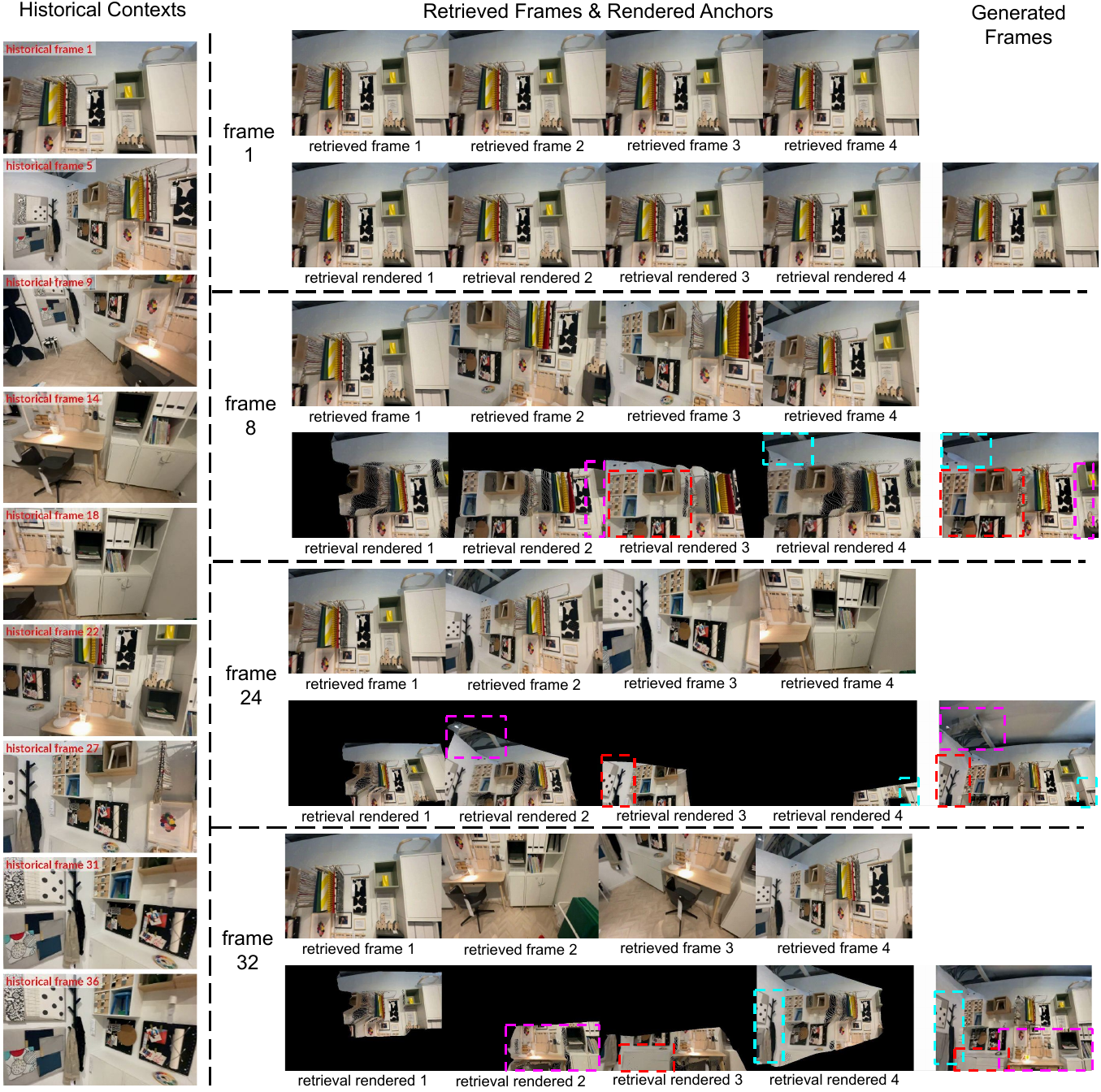}
\caption{Visualization of AnchorWeave’s retrieval → rendering → generation process.
 We show representative historical contexts (serving as memory pool),  retrieved frames with retrieval rendered anchors, and final generations. For each target frame, all retrievals' anchors provide complementary geometric evidence, and AnchorWeave reconciles them into a coherent output.
Same-colored dashed boxes highlight preserved regions.
    }
    \label{fig:main vis}
\end{figure*}

\begin{figure*}[!t]
    \centering
    \includegraphics[width=\linewidth]{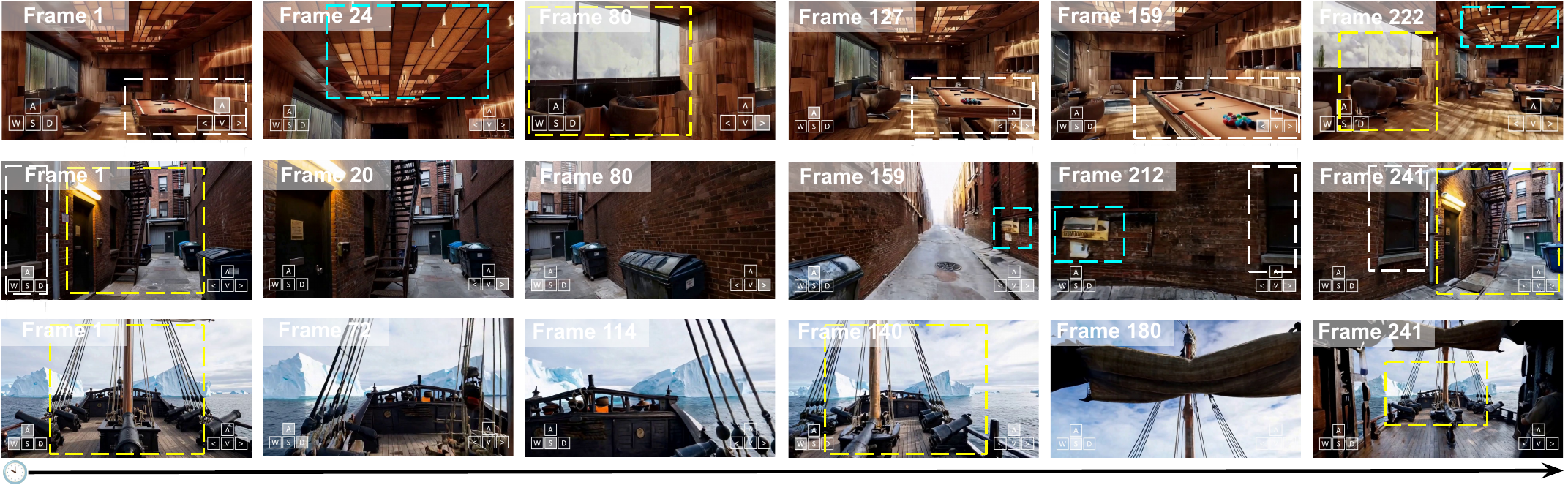}
\caption{Long-horizon world exploration examples with keyboard-controlled camera actions on open-domain images.
AnchorWeave preserves consistent object geometry and appearance across frames (highlighted by dashed boxes of the same color).
}
    \label{fig:open-domain}
\end{figure*}

\myparagraph{Effect of joint vs.\ separate attention in the controller.}
\Cref{fig:ablation}(b) compares the proposed joint-attention-based controller with a baseline that processes each retrieved anchor independently using separate attention blocks.
We observe that separate attention leads to blurred or inconsistent structures due to the absence of cross-anchor information exchange.
In contrast, joint attention allows the model to aggregate complementary evidence across anchors, resulting in sharper and more coherent geometry in generated frames.

\myparagraph{Effect of the number of retrieved frames.}
Table~\ref{tab:ablation_num_retrievals} studies the impact of the number of retrieved frames by varying the number of retrievals $K$ per chunk.
Increasing the number of retrieved frames consistently improves PSNR and SSIM, indicating that additional anchors provide complementary spatial evidence for generation.
This trend is also reflected in the comparisons in~\Cref{fig:qual_com}, where more anchors leads to reduced misalignment and improved consistency.

\myparagraph{Efficiency and cost.}
\Cref{tab:efficiency_main} summarizes runtime on an RTX PRO 6000 GPU with Wan2.2-5B at $81{\times}480{\times}832$ resolution. AnchorWeave adds moderate generation overhead over SPMem due to multi-anchor control ($275$ vs.\ $221$s; $39$ vs.\ $34$GB at $K=4$), but this comes with a large quality gain over SPMem in \Cref{tab:main_re10k_mira}. Increasing $K$ gives a predictable trade-off: moving from $K=1$ to $K=4$ improves PSNR by roughly 2dB while adding 54s and 5GB. A full history-length scaling analysis is provided in the Appendix, showing that local-memory retrieval remains more stable as history grows, whereas SPMem's accumulated global-point-cloud rendering cost grows faster.

\begin{table}[t]
\centering

\caption{Efficiency and performance--cost trade-off. Runtime is measured on an RTX PRO 6000 GPU with Wan2.2-5B at $81{\times}480{\times}832$.}
\label{tab:efficiency_main}
\scriptsize
\setlength{\tabcolsep}{3.5pt}
\begin{tabular}{lcc}
\toprule
Component / setting & Time (s) & GPU (GB) \\
\midrule
TTT3R per-frame memory & 0.06 & -- \\
Retrieval + rendering (ours) & avg. 15 & -- \\
Retrieval + rendering (SPMem) & avg. 13 & -- \\
Video generation (SPMem) & 221 & 34 \\
Video generation (ours, $K=4$) & 275 & 39 \\
\midrule
$K=1$ (PSNR 19.0) & 221 & 34 \\
$K=2$ (PSNR 20.0) & 238 & 35 \\
$K=4$ (PSNR 21.0) & 275 & 39 \\
\bottomrule
\end{tabular}

\end{table}

\subsection{AnchorWeave Process Visualization}

We visualize the AnchorWeave pipeline (retrieve $\rightarrow$ render $\rightarrow$ generate) in \Cref{fig:main vis}.
For each target frame, multiple coverage-relevant anchors are retrieved and rendered under the target view. Following our implementation details (\Cref{implementation}), the first rendered frame is always selected as the initial anchor, serving as a stable geometric reference.
However, under viewpoint changes, it cannot cover all regions previously observed in the historical contexts.
Subsequent retrievals (anchors 2–4), selected via coverage-driven retrieval, complement the initial anchor by providing geometric evidence for the regions where the first frame doesn't cover.
Although individual anchors may exhibit misalignment, the joint multi-anchor controller weaves these complementary signals during generation, merging evidence from different anchors into a coherent frame.
As a result, information from regions unseen in the first anchor is consistently integrated, preserving strong spatial consistency, as shown in the highlighted dashed boxes with the same colors.
Additional examples are provided in the Appendix.

\subsection{Open-Domain Long-Horizon Examples}
We present three open-domain long-horizon exploration examples in~\Cref{fig:open-domain} using AnchorWeave with Wan2.2 backbone. 
The model generalizes effectively to diverse unseen scenarios, such as a wooden cabin, an alley, and a wooden sailing ship (top, middle, and bottom rows, respectively).
While the visual backbone is limited to 81-frame video, AnchorWeave enables generation beyond this limit by conditioning each new segment on accumulated memory. 
As a result, spatial consistency is preserved across multiple segments (e.g. in the cabin scene, regions observed at frame 1 and in later frames (24 and 80), such as the ceiling and stools, remain consistent in the final frame (frame 222);
in the alley scene, 360° panoramic consistency is preserved, with strong alignment between the initial and terminal frames). More long-horizon world exploration examples are provided in the Appendix, highlighting long-term consistency, 360° panoramic scene generation, and generalization to third-person gaming scenarios.

\section{Conclusion}
We present AnchorWeave, a geometry-informed, memory-aware video generation framework that improves long-horizon generation by replacing brittle global 3D reconstruction with multiple clean, view-aligned local geometric memories. Through coverage-driven retrieval and adaptive multi-anchor fusion, AnchorWeave provides robust spatial guidance and significantly improves visual quality, camera controllability, and long-term consistency across diverse scenes and tasks. Our results show that conditioning on local geometric memories and learning to reconcile their inconsistencies offers a reliable alternative to global geometric fusion for scalable long-horizon video generation.

\section*{Acknowledgments}

This work was supported by DARPA ECOLE Program No. HR00112390060, NSF-AI Engage Institute DRL-2112635, DARPA Machine Commonsense (MCS) Grant N66001-19-2-4031, ARO Award W911NF2110220, ONR Grant N00014-23-1-2356, Accelerate Foundation Models Research program, and a Bloomberg Data Science PhD Fellowship. The views contained in this article are those of the authors and not of the funding agency.

\clearpage
{
\small
\bibliographystyle{splncs04}
\bibliography{main}

@String(CVPR  = {IEEE Conf. Comput. Vis. Pattern Recog.})

@String(NeurIPS = {Adv. Neural Inform. Process. Syst.})

@String(ICLR  = {Int. Conf. Learn. Represent.})

@String(TOG   = {ACM Trans. Graph.})

@String(CVPR  = {CVPR})

@String(NeurIPS = {NeurIPS})

@String(ICLR  = {ICLR})

@String(TOG   = {ACM TOG})

@article{gu2025long,
  title={Long-context autoregressive video modeling with next-frame prediction},
  author={Gu, Yuchao and Mao, Weijia and Shou, Mike Zheng},
  journal={arXiv preprint arXiv:2503.19325},
  year={2025}
}

@article{guo2024infinitydrive,
  title={Infinitydrive: Breaking time limits in driving world models},
  author={Guo, Xi and Ding, Chenjing and Dou, Haoxuan and Zhang, Xin and Tang, Weixuan and Wu, Wei},
  journal={arXiv preprint arXiv:2412.01522},
  year={2024}
}

@article{he2024cameractrl,
  title={Cameractrl: Enabling camera control for text-to-video generation},
  author={He, Hao and Xu, Yinghao and Guo, Yuwei and Wetzstein, Gordon and Dai, Bo and Li, Hongsheng and Yang, Ceyuan},
  journal={arXiv preprint arXiv:2404.02101},
  year={2024}
}

@article{wang2025epic,
  title={EPiC: Efficient Video Camera Control Learning with Precise Anchor-Video Guidance},
  author={Wang, Zun and Cho, Jaemin and Li, Jialu and Lin, Han and Yoon, Jaehong and Zhang, Yue and Bansal, Mohit},
  journal={arXiv preprint arXiv:2505.21876},
  year={2025}
}

@article{hou2024learning,
  title={Learning camera movement control from real-world drone videos},
  author={Hou, Yunzhong and Zheng, Liang and Torr, Philip},
  journal={arXiv preprint arXiv:2412.09620},
  year={2024}
}

@inproceedings{yu2025context,
  title={Context as memory: Scene-consistent interactive long video generation with memory retrieval},
  author={Yu, Jiwen and Bai, Jianhong and Qin, Yiran and Liu, Quande and Wang, Xintao and Wan, Pengfei and Zhang, Di and Liu, Xihui},
  booktitle={Proceedings of the SIGGRAPH Asia 2025 Conference Papers},
  pages={1--11},
  year={2025}
}

@article{li2025vmem,
  title={VMem: Consistent Interactive Video Scene Generation with Surfel-Indexed View Memory},
  author={Li, Runjia and Torr, Philip and Vedaldi, Andrea and Jakab, Tomas},
  journal={arXiv preprint arXiv:2506.18903},
  year={2025}
}

@article{xiao2025worldmem,
  title={Worldmem: Long-term consistent world simulation with memory},
  author={Xiao, Zeqi and Lan, Yushi and Zhou, Yifan and Ouyang, Wenqi and Yang, Shuai and Zeng, Yanhong and Pan, Xingang},
  journal={arXiv preprint arXiv:2504.12369},
  year={2025}
}

@misc{wu2025spmem,
      title={Video World Models with Long-term Spatial Memory}, 
      author={Tong Wu and Shuai Yang and Ryan Po and Yinghao Xu and Ziwei Liu and Dahua Lin and Gordon Wetzstein},
      year={2025},
      eprint={2506.05284},
      archivePrefix={arXiv},
      primaryClass={cs.CV},
      url={https://arxiv.org/abs/2506.05284}, 
}

@article{zhao2025spatia,
  title={Spatia: Video Generation with Updatable Spatial Memory},
  author={Zhao, Jinjing and Wei, Fangyun and Liu, Zhening and Zhang, Hongyang and Xu, Chang and Lu, Yan},
  journal={arXiv preprint arXiv:2512.15716},
  year={2025}
}

@article{li2025magicworld,
  title={MagicWorld: Interactive Geometry-driven Video World Exploration},
  author={Li, Guangyuan and Zheng, Siming and Xu, Shuolin and Chen, Jinwei and Li, Bo and Hu, Xiaobin and Zhao, Lei and Jiang, Peng-Tao},
  journal={arXiv preprint arXiv:2511.18886},
  year={2025}
}

@article{yang2024cogvideox,
  title={CogVideoX: Text-to-Video Diffusion Models with An Expert Transformer},
  author={Yang, Zhuoyi and Teng, Jiayan and Zheng, Wendi and Ding, Ming and Huang, Shiyu and Xu, Jiazheng and Yang, Yuanming and Hong, Wenyi and Zhang, Xiaohan and Feng, Guanyu and others},
  journal={arXiv preprint arXiv:2408.06072},
  year={2024}
}

@article{chen2025ttt3r,
    title={TTT3R: 3D Reconstruction as Test-Time Training},
    author={Chen, Xingyu and Chen, Yue and Xiu, Yuliang and Geiger, Andreas and Chen, Anpei},
    journal={arXiv preprint arXiv:2509.26645},
    year={2025}
    }

@article{RelicWorldModel2025,
  title={RELIC: Interactive Video World Model with Long-Horizon Memory},
  author={Hong, Yicong and Mei, Yiqun and Ge, Chongjian and Xu, Yiran and Zhou, Yang and Bi, Sai and Hold-Geoffroy, Yannick and Roberts, Mike and Fisher, Matthew and Shechtman, Eli and others},
  journal={arXiv preprint arXiv:2512.04040},
  year={2025}
}

@article{sun2025worldplay,
  title={Worldplay: Towards long-term geometric consistency for real-time interactive world modeling},
  author={Sun, Wenqiang and Zhang, Haiyu and Wang, Haoyuan and Wu, Junta and Wang, Zehan and Wang, Zhenwei and Wang, Yunhong and Zhang, Jun and Wang, Tengfei and Guo, Chunchao},
  journal={arXiv preprint arXiv:2512.14614},
  year={2025}
}

@inproceedings{rombach2022high,
  title={High-resolution image synthesis with latent diffusion models},
  author={Rombach, Robin and Blattmann, Andreas and Lorenz, Dominik and Esser, Patrick and Ommer, Bj{\"o}rn},
  booktitle={Proceedings of the IEEE/CVF conference on computer vision and pattern recognition},
  pages={10684--10695},
  year={2022}
}

@article{ho2020denoising,
  title={Denoising diffusion probabilistic models},
  author={Ho, Jonathan and Jain, Ajay and Abbeel, Pieter},
  journal={Advances in neural information processing systems},
  volume={33},
  pages={6840--6851},
  year={2020}
}

@inproceedings{lipmanflow,
  title={Flow Matching for Generative Modeling},
  author={Lipman, Yaron and Chen, Ricky TQ and Ben-Hamu, Heli and Nickel, Maximilian and Le, Matthew},
  booktitle={The Eleventh International Conference on Learning Representations},
year={2023}
}

@inproceedings{bruce2024genie,
  title={Genie: Generative interactive environments},
  author={Bruce, Jake and Dennis, Michael D and Edwards, Ashley and Parker-Holder, Jack and Shi, Yuge and Hughes, Edward and Lai, Matthew and Mavalankar, Aditi and Steigerwald, Richie and Apps, Chris and others},
  booktitle={Forty-first International Conference on Machine Learning},
  year={2024}
}

@article{parkerholder2024genie2,
  title         = {Genie 2: A Large-Scale Foundation World Model},
  author        = {Jack Parker-Holder and Philip Ball and Jake Bruce and Vibhavari Dasagi and Kristian Holsheimer and Christos Kaplanis and Alexandre Moufarek and Guy Scully and Jeremy Shar and Jimmy Shi and Stephen Spencer and Jessica Yung and Michael Dennis and Sultan Kenjeyev and Shangbang Long and Vlad Mnih and Harris Chan and Maxime Gazeau and Bonnie Li and Fabio Pardo and Luyu Wang and Lei Zhang and Frederic Besse and Tim Harley and Anna Mitenkova and Jane Wang and Jeff Clune and Demis Hassabis and Raia Hadsell and Adrian Bolton and Satinder Singh and Tim Rockt{\"a}schel},
  journal       = {Google DeepMind},
  year          = {2024},
  url           = {https://deepmind.google/discover/blog/genie-2-a-large-scale-foundation-world-model/}
}

@article{genie3,
  title         = {Genie 3: A New Frontier for World Models},
  author        = {Philip J. Ball and Jakob Bauer and Frank Belletti and Bethanie Brownfield and Ariel Ephrat and Shlomi Fruchter and Agrim Gupta and Kristian Holsheimer and Aleksander Holynski and Jiri Hron and Christos Kaplanis and Marjorie Limont and Matt McGill and Yanko Oliveira and Jack Parker-Holder and Frank Perbet and Guy Scully and Jeremy Shar and Stephen Spencer and Omer Tov and Ruben Villegas and Emma Wang and Jessica Yung and Cip Baetu and Jordi Berbel and David Bridson and Jake Bruce and Gavin Buttimore and Sarah Chakera and Bilva Chandra and Paul Collins and Alex Cullum and Bogdan Damoc and Vibha Dasagi and Maxime Gazeau and Charles Gbadamosi and Woohyun Han and Ed Hirst and Ashyana Kachra and Lucie Kerley and Kristian Kjems and Eva Knoepfel and Vika Koriakin and Jessica Lo and Cong Lu and Zeb Mehring and Alex Moufarek and Henna Nandwani and Valeria Oliveira and Fabio Pardo and Jane Park and Andrew Pierson and Ben Poole and Helen Ran and Tim Salimans and Manuel Sanchez and Igor Saprykin and Amy Shen and Sailesh Sidhwani and Duncan Smith and Joe Stanton and Hamish Tomlinson and Dimple Vijaykumar and Luyu Wang and Piers Wingfield and Nat Wong and Keyang Xu and Christopher Yew and Nick Young and Vadim Zubov and Douglas Eck and Dumitru Erhan and Koray Kavukcuoglu and Demis Hassabis and Zoubin Gharamani and Raia Hadsell and A{\"a}ron van den Oord and Inbar Mosseri and Adrian Bolton and Satinder Singh and Tim Rockt{\"a}schel},
  journal       = {Google DeepMind},
  year          = {2025},
  url           = {https://deepmind.google/blog/genie-3-a-new-frontier-for-world-models/}
}

@article{guo2025ctrl,
  title={Ctrl-world: A controllable generative world model for robot manipulation},
  author={Guo, Yanjiang and Shi, Lucy Xiaoyang and Chen, Jianyu and Finn, Chelsea},
  journal={arXiv preprint arXiv:2510.10125},
  year={2025}
}

@article{chen2025learning,
  title={Learning World Models for Interactive Video Generation},
  author={Chen, Taiye and Hu, Xun and Ding, Zihan and Jin, Chi},
  journal={arXiv preprint arXiv:2505.21996},
  year={2025}
}

@article{wan2025,
      title={Wan: Open and Advanced Large-Scale Video Generative Models}, 
      author={Team Wan and Ang Wang and Baole Ai and Bin Wen and Chaojie Mao and Chen-Wei Xie and Di Chen and Feiwu Yu and Haiming Zhao and Jianxiao Yang and Jianyuan Zeng and Jiayu Wang and Jingfeng Zhang and Jingren Zhou and Jinkai Wang and Jixuan Chen and Kai Zhu and Kang Zhao and Keyu Yan and Lianghua Huang and Mengyang Feng and Ningyi Zhang and Pandeng Li and Pingyu Wu and Ruihang Chu and Ruili Feng and Shiwei Zhang and Siyang Sun and Tao Fang and Tianxing Wang and Tianyi Gui and Tingyu Weng and Tong Shen and Wei Lin and Wei Wang and Wei Wang and Wenmeng Zhou and Wente Wang and Wenting Shen and Wenyuan Yu and Xianzhong Shi and Xiaoming Huang and Xin Xu and Yan Kou and Yangyu Lv and Yifei Li and Yijing Liu and Yiming Wang and Yingya Zhang and Yitong Huang and Yong Li and You Wu and Yu Liu and Yulin Pan and Yun Zheng and Yuntao Hong and Yupeng Shi and Yutong Feng and Zeyinzi Jiang and Zhen Han and Zhi-Fan Wu and Ziyu Liu},
      journal = {arXiv preprint arXiv:2503.20314},
      year={2025}
}

@article{kong2024hunyuanvideo,
  title={Hunyuanvideo: A systematic framework for large video generative models},
  author={Kong, Weijie and Tian, Qi and Zhang, Zijian and Min, Rox and Dai, Zuozhuo and Zhou, Jin and Xiong, Jiangfeng and Li, Xin and Wu, Bo and Zhang, Jianwei and others},
  journal={arXiv preprint arXiv:2412.03603},
  year={2024}
}

@article{hacohen2026ltx,
  title={LTX-2: Efficient Joint Audio-Visual Foundation Model},
  author={HaCohen, Yoav and Brazowski, Benny and Chiprut, Nisan and Bitterman, Yaki and Kvochko, Andrew and Berkowitz, Avishai and Shalem, Daniel and Lifschitz, Daphna and Moshe, Dudu and Porat, Eitan and others},
  journal={arXiv preprint arXiv:2601.03233},
  year={2026}
}

@article{huang2025selfforcing,
  title={Self Forcing: Bridging the Train-Test Gap in Autoregressive Video Diffusion},
  author={Huang, Xun and Li, Zhengqi and He, Guande and Zhou, Mingyuan and Shechtman, Eli},
  journal={arXiv preprint arXiv:2506.08009},
  year={2025}
}

@article{po2025long,
  title={Long-context state-space video world models},
  author={Po, Ryan and Nitzan, Yotam and Zhang, Richard and Chen, Berlin and Dao, Tri and Shechtman, Eli and Wetzstein, Gordon and Huang, Xun},
  journal={arXiv preprint arXiv:2505.20171},
  year={2025}
}

@article{savov2025statespacediffuser,
  title={StateSpaceDiffuser: Bringing Long Context to Diffusion World Models},
  author={Savov, Nedko and Kazemi, Naser and Zhang, Deheng and Paudel, Danda Pani and Wang, Xi and Van Gool, Luc},
  journal={arXiv preprint arXiv:2505.22246},
  year={2025}
}

@article{lee2025enhancing,
  title={Enhancing Memory and Imagination Consistency in Diffusion-based World Models via Linear-Time Sequence Modeling},
  author={Lee, Jia-Hua and Lin, Bor-Jiun and Sun, Wei-Fang and Lee, Chun-Yi},
  journal={arXiv e-prints},
  pages={arXiv--2502},
  year={2025}
}

@article{he2025matrix,
  title={Matrix-game 2.0: An open-source real-time and streaming interactive world model},
  author={He, Xianglong and Peng, Chunli and Liu, Zexiang and Wang, Boyang and Zhang, Yifan and Cui, Qi and Kang, Fei and Jiang, Biao and An, Mengyin and Ren, Yangyang and others},
  journal={arXiv preprint arXiv:2508.13009},
  year={2025}
}

@article{huang2025self,
  title={Self Forcing: Bridging the Train-Test Gap in Autoregressive Video Diffusion},
  author={Huang, Xun and Li, Zhengqi and He, Guande and Zhou, Mingyuan and Shechtman, Eli},
  journal={arXiv preprint arXiv:2506.08009},
  year={2025}
}

@article{yang2025longlive,
  title={Longlive: Real-time interactive long video generation},
  author={Yang, Shuai and Huang, Wei and Chu, Ruihang and Xiao, Yicheng and Zhao, Yuyang and Wang, Xianbang and Li, Muyang and Xie, Enze and Chen, Yingcong and Lu, Yao and others},
  journal={arXiv preprint arXiv:2509.22622},
  year={2025}
}

@inproceedings{zhang2025frame,
  title={Frame context packing and drift prevention in next-frame-prediction video diffusion models},
  author={Zhang, Lvmin and Cai, Shengqu and Li, Muyang and Wetzstein, Gordon and Agrawala, Maneesh},
  booktitle={The Thirty-ninth Annual Conference on Neural Information Processing Systems},
  year={2025}
}

@article{mao2025yume,
  title={Yume-1.5: A Text-Controlled Interactive World Generation Model},
  author={Mao, Xiaofeng and Li, Zhen and Li, Chuanhao and Xu, Xiaojie and Ying, Kaining and He, Tong and Pang, Jiangmiao and Qiao, Yu and Zhang, Kaipeng},
  journal={arXiv preprint arXiv:2512.22096},
  year={2025}
}

@article{ye2025yan,
  title={Yan: Foundational interactive video generation},
  author={Ye, Deheng and Zhou, Fangyun and Lv, Jiacheng and Ma, Jianqi and Zhang, Jun and Lv, Junyan and Li, Junyou and Deng, Minwen and Yang, Mingyu and Fu, Qiang and others},
  journal={arXiv preprint arXiv:2508.08601},
  year={2025}
}

@article{liu2025rolling,
  title={Rolling forcing: Autoregressive long video diffusion in real time},
  author={Liu, Kunhao and Hu, Wenbo and Xu, Jiale and Shan, Ying and Lu, Shijian},
  journal={arXiv preprint arXiv:2509.25161},
  year={2025}
}

@article{zhou2025learning,
  title={Learning 3D Persistent Embodied World Models},
  author={Zhou, Siyuan and Du, Yilun and Yang, Yuncong and Han, Lei and Chen, Peihao and Yeung, Dit-Yan and Gan, Chuang},
  journal={arXiv preprint arXiv:2505.05495},
  year={2025}
}

@inproceedings{liu2025dynamem,
  title={Dynamem: Online dynamic spatio-semantic memory for open world mobile manipulation},
  author={Liu, Peiqi and Guo, Zhanqiu and Warke, Mohit and Chintala, Soumith and Paxton, Chris and Shafiullah, Nur Muhammad Mahi and Pinto, Lerrel},
  booktitle={2025 IEEE International Conference on Robotics and Automation (ICRA)},
  pages={13346--13355},
  year={2025},
  organization={IEEE}
}

@article{zhang2025test,
  title={Test-time training done right},
  author={Zhang, Tianyuan and Bi, Sai and Hong, Yicong and Zhang, Kai and Luan, Fujun and Yang, Songlin and Sunkavalli, Kalyan and Freeman, William T and Tan, Hao},
  journal={arXiv preprint arXiv:2505.23884},
  year={2025}
}

@inproceedings{dalal2025one,
  title={One-minute video generation with test-time training},
  author={Dalal, Karan and Koceja, Daniel and Xu, Jiarui and Zhao, Yue and Han, Shihao and Cheung, Ka Chun and Kautz, Jan and Choi, Yejin and Sun, Yu and Wang, Xiaolong},
  booktitle={Proceedings of the Computer Vision and Pattern Recognition Conference},
  pages={17702--17711},
  year={2025}
}

@article{hong2024slowfast,
  title={Slowfast-vgen: Slow-fast learning for action-driven long video generation},
  author={Hong, Yining and Liu, Beide and Wu, Maxine and Zhai, Yuanhao and Chang, Kai-Wei and Li, Linjie and Lin, Kevin and Lin, Chung-Ching and Wang, Jianfeng and Yang, Zhengyuan and others},
  journal={arXiv preprint arXiv:2410.23277},
  year={2024}
}

@inproceedings{chendiffusion,
  title={Diffusion Forcing: Next-token Prediction Meets Full-Sequence Diffusion},
  author={Chen, Boyuan and Mons{\'o}, Diego Mart{\'\i} and Du, Yilun and Simchowitz, Max and Tedrake, Russ and Sitzmann, Vincent},
  booktitle={The Thirty-eighth Annual Conference on Neural Information Processing Systems},
year={2025}
}

@inproceedings{ling2024dl3dv,
  title={Dl3dv-10k: A large-scale scene dataset for deep learning-based 3d vision},
  author={Ling, Lu and Sheng, Yichen and Tu, Zhi and Zhao, Wentian and Xin, Cheng and Wan, Kun and Yu, Lantao and Guo, Qianyu and Yu, Zixun and Lu, Yawen and others},
  booktitle={Proceedings of the IEEE/CVF Conference on Computer Vision and Pattern Recognition},
  pages={22160--22169},
  year={2024}
}

@inproceedings{ren2025gen3c,
  title={Gen3c: 3d-informed world-consistent video generation with precise camera control},
  author={Ren, Xuanchi and Shen, Tianchang and Huang, Jiahui and Ling, Huan and Lu, Yifan and Nimier-David, Merlin and M{\"u}ller, Thomas and Keller, Alexander and Fidler, Sanja and Gao, Jun},
  booktitle={Proceedings of the Computer Vision and Pattern Recognition Conference},
  pages={6121--6132},
  year={2025}
}

@article{yu2025trajectorycrafter,
  title={Trajectorycrafter: Redirecting camera trajectory for monocular videos via diffusion models},
  author={YU, Mark and Hu, Wenbo and Xing, Jinbo and Shan, Ying},
  journal={arXiv preprint arXiv:2503.05638},
  year={2025}
}

@article{yu2024viewcrafter,
  title={Viewcrafter: Taming video diffusion models for high-fidelity novel view synthesis},
  author={Yu, Wangbo and Xing, Jinbo and Yuan, Li and Hu, Wenbo and Li, Xiaoyu and Huang, Zhipeng and Gao, Xiangjun and Wong, Tien-Tsin and Shan, Ying and Tian, Yonghong},
  journal={arXiv preprint arXiv:2409.02048},
  year={2024}
}

@inproceedings{wang2024motionctrl,
  title={Motionctrl: A unified and flexible motion controller for video generation},
  author={Wang, Zhouxia and Yuan, Ziyang and Wang, Xintao and Li, Yaowei and Chen, Tianshui and Xia, Menghan and Luo, Ping and Shan, Ying},
  booktitle={ACM SIGGRAPH 2024 Conference Papers},
  pages={1--11},
  year={2024}
}

@inproceedings{watson2024controlling,
  title={Controlling space and time with diffusion models},
  author={Watson, Daniel and Saxena, Saurabh and Li, Lala and Tagliasacchi, Andrea and Fleet, David J},
  booktitle={The Thirteenth International Conference on Learning Representations},
  year={2024}
}

@inproceedings{
yuegosim,
title={EgoSim: Egocentric Exploration in Virtual Worlds with Multi-modal Conditioning},
author={Wei Yu and Songheng Yin and Steve Easterbrook and Animesh Garg},
booktitle={The Thirteenth International Conference on Learning Representations},
year={2025},
url={https://openreview.net/forum?id=zAyS5aRKV8}
}

@article{bahmani2024ac3d,
  title={AC3D: Analyzing and Improving 3D Camera Control in Video Diffusion Transformers},
  author={Bahmani, Sherwin and Skorokhodov, Ivan and Qian, Guocheng and Siarohin, Aliaksandr and Menapace, Willi and Tagliasacchi, Andrea and Lindell, David B and Tulyakov, Sergey},
  journal={arXiv preprint arXiv:2411.18673},
  year={2024}
}

@article{bahmani2024vd3d,
  title={Vd3d: Taming large video diffusion transformers for 3d camera control},
  author={Bahmani, Sherwin and Skorokhodov, Ivan and Siarohin, Aliaksandr and Menapace, Willi and Qian, Guocheng and Vasilkovsky, Michael and Lee, Hsin-Ying and Wang, Chaoyang and Zou, Jiaxu and Tagliasacchi, Andrea and others},
  journal={arXiv preprint arXiv:2407.12781},
  year={2024}
}

@article{li2025realcam,
  title={Realcam-i2v: Real-world image-to-video generation with interactive complex camera control},
  author={Li, Teng and Zheng, Guangcong and Jiang, Rui and Wu, Tao and Lu, Yehao and Lin, Yining and Li, Xi and others},
  journal={arXiv preprint arXiv:2502.10059},
  year={2025}
}

@article{zheng2024cami2v,
  title={Cami2v: Camera-controlled image-to-video diffusion model},
  author={Zheng, Guangcong and Li, Teng and Jiang, Rui and Lu, Yehao and Wu, Tao and Li, Xi},
  journal={arXiv preprint arXiv:2410.15957},
  year={2024}
}

@article{sun2024dimensionx,
  title={Dimensionx: Create any 3d and 4d scenes from a single image with controllable video diffusion},
  author={Sun, Wenqiang and Chen, Shuo and Liu, Fangfu and Chen, Zilong and Duan, Yueqi and Zhang, Jun and Wang, Yikai},
  journal={arXiv preprint arXiv:2411.04928},
  year={2024}
}

@article{he2025cameractrl,
  title={CameraCtrl II: Dynamic Scene Exploration via Camera-controlled Video Diffusion Models},
  author={He, Hao and Yang, Ceyuan and Lin, Shanchuan and Xu, Yinghao and Wei, Meng and Gui, Liangke and Zhao, Qi and Wetzstein, Gordon and Jiang, Lu and Li, Hongsheng},
  journal={arXiv preprint arXiv:2503.10592},
  year={2025}
}

@article{zhou2025stable,
  title={STABLE VIRTUAL CAMERA: Generative View Synthesis with Diffusion Models},
  author={Zhou, Jensen Jinghao and Gao, Hang and Voleti, Vikram and Vasishta, Aaryaman and Yao, Chun-Han and Boss, Mark and Torr, Philip and Rupprecht, Christian and Jampani, Varun},
  journal={arXiv e-prints},
  pages={arXiv--2503},
  year={2025}
}

@inproceedings{van2024generative,
  title={Generative camera dolly: Extreme monocular dynamic novel view synthesis},
  author={Van Hoorick, Basile and Wu, Rundi and Ozguroglu, Ege and Sargent, Kyle and Liu, Ruoshi and Tokmakov, Pavel and Dave, Achal and Zheng, Changxi and Vondrick, Carl},
  booktitle={European Conference on Computer Vision},
  pages={313--331},
  year={2024},
  organization={Springer}
}

@article{jin2025flovd,
  title={FloVD: Optical Flow Meets Video Diffusion Model for Enhanced Camera-Controlled Video Synthesis},
  author={Jin, Wonjoon and Dai, Qi and Luo, Chong and Baek, Seung-Hwan and Cho, Sunghyun},
  journal={arXiv preprint arXiv:2502.08244},
  year={2025}
}

@article{bai2025recammaster,
  title={ReCamMaster: Camera-Controlled Generative Rendering from A Single Video},
  author={Bai, Jianhong and Xia, Menghan and Fu, Xiao and Wang, Xintao and Mu, Lianrui and Cao, Jinwen and Liu, Zuozhu and Hu, Haoji and Bai, Xiang and Wan, Pengfei and others},
  journal={arXiv preprint arXiv:2503.11647},
  year={2025}
}

@article{cao2025uni3c,
  title={Uni3C: Unifying Precisely 3D-Enhanced Camera and Human Motion Controls for Video Generation},
  author={Cao, Chenjie and Zhou, Jingkai and Li, Shikai and Liang, Jingyun and Yu, Chaohui and Wang, Fan and Xue, Xiangyang and Fu, Yanwei},
  journal={arXiv preprint arXiv:2504.14899},
  year={2025}
}

@article{zhou2018stereo,
  title={Stereo magnification: Learning view synthesis using multiplane images},
  author={Zhou, Tinghui and Tucker, Richard and Flynn, John and Fyffe, Graham and Snavely, Noah},
  journal={arXiv preprint arXiv:1805.09817},
  year={2018}
}

@article{you2024nvs,
  title={Nvs-solver: Video diffusion model as zero-shot novel view synthesizer},
  author={You, Meng and Zhu, Zhiyu and Liu, Hui and Hou, Junhui},
  journal={arXiv preprint arXiv:2405.15364},
  year={2024}
}

@article{zhou2024latent,
  title={Latent-Reframe: Enabling Camera Control for Video Diffusion Model without Training},
  author={Zhou, Zhenghong and An, Jie and Luo, Jiebo},
  journal={arXiv preprint arXiv:2412.06029},
  year={2024}
}

@article{popov2025camctrl3d,
  title={CamCtrl3D: Single-Image Scene Exploration with Precise 3D Camera Control},
  author={Popov, Stefan and Raj, Amit and Krainin, Michael and Li, Yuanzhen and Freeman, William T and Rubinstein, Michael},
  journal={arXiv preprint arXiv:2501.06006},
  year={2025}
}

@article{hou2024training,
  title={Training-free camera control for video generation},
  author={Hou, Chen and Wei, Guoqiang and Zeng, Yan and Chen, Zhibo},
  journal={arXiv preprint arXiv:2406.10126},
  year={2024}
}

@inproceedings{hecameractrl,
  title={CameraCtrl: Enabling Camera Control for Video Diffusion Models},
  author={He, Hao and Xu, Yinghao and Guo, Yuwei and Wetzstein, Gordon and Dai, Bo and Li, Hongsheng and Yang, Ceyuan},
  booktitle={The Thirteenth International Conference on Learning Representations},
  year={2025}
}

@article{li2024nvcomposer,
  title={Nvcomposer: Boosting generative novel view synthesis with multiple sparse and unposed images},
  author={Li, Lingen and Zhang, Zhaoyang and Li, Yaowei and Xu, Jiale and Hu, Wenbo and Li, Xiaoyu and Cheng, Weihao and Gu, Jinwei and Xue, Tianfan and Shan, Ying},
  journal={arXiv preprint arXiv:2412.03517},
  year={2024}
}

@article{zheng2025vidcraft3,
  title={VidCRAFT3: Camera, Object, and Lighting Control for Image-to-Video Generation},
  author={Zheng, Sixiao and Peng, Zimian and Zhou, Yanpeng and Zhu, Yi and Xu, Hang and Huang, Xiangru and Fu, Yanwei},
  journal={arXiv preprint arXiv:2502.07531},
  year={2025}
}

@inproceedings{seo2024genwarp,
  title={Genwarp: Single image to novel views with semantic-preserving generative warping},
  author={Seo, Junyoung and Fukuda, Kazumi and Shibuya, Takashi and Narihira, Takuya and Murata, Naoki and Hu, Shoukang and Lai, Chieh-Hsin and Kim, Seungryong and Mitsufuji, Yuki},
  booktitle={The Thirty-eighth Annual Conference on Neural Information Processing Systems},
  year={2024}
}

@article{zhang2025i2v3d,
  title={I2V3D: Controllable image-to-video generation with 3D guidance},
  author={Zhang, Zhiyuan and Chen, Dongdong and Liao, Jing},
  journal={arXiv preprint arXiv:2503.09733},
  year={2025}
}

@article{feng2024i2vcontrol,
  title={I2VControl-Camera: Precise Video Camera Control with Adjustable Motion Strength},
  author={Feng, Wanquan and Liu, Jiawei and Tu, Pengqi and Qi, Tianhao and Sun, Mingzhen and Ma, Tianxiang and Zhao, Songtao and Zhou, Siyu and He, Qian},
  journal={arXiv preprint arXiv:2411.06525},
  year={2024}
}

@article{xiao2024trajectory,
  title={Trajectory Attention for Fine-grained Video Motion Control},
  author={Xiao, Zeqi and Ouyang, Wenqi and Zhou, Yifan and Yang, Shuai and Yang, Lei and Si, Jianlou and Pan, Xingang},
  journal={arXiv preprint arXiv:2411.19324},
  year={2024}
}

@article{gu2025diffusion,
  title={Diffusion as Shader: 3D-aware Video Diffusion for Versatile Video Generation Control},
  author={Gu, Zekai and Yan, Rui and Lu, Jiahao and Li, Peng and Dou, Zhiyang and Si, Chenyang and Dong, Zhen and Liu, Qifeng and Lin, Cheng and Liu, Ziwei and others},
  journal={arXiv preprint arXiv:2501.03847},
  year={2025}
}

@inproceedings{greff2022kubric,
  title={Kubric: A scalable dataset generator},
  author={Greff, Klaus and Belletti, Francois and Beyer, Lucas and Doersch, Carl and Du, Yilun and Duckworth, Daniel and Fleet, David J and Gnanapragasam, Dan and Golemo, Florian and Herrmann, Charles and others},
  booktitle={Proceedings of the IEEE/CVF conference on computer vision and pattern recognition},
  pages={3749--3761},
  year={2022}
}

@article{chen2024diffusion,
  title={Diffusion forcing: Next-token prediction meets full-sequence diffusion},
  author={Chen, Boyuan and Mart{\'\i} Mons{\'o}, Diego and Du, Yilun and Simchowitz, Max and Tedrake, Russ and Sitzmann, Vincent},
  journal={Advances in Neural Information Processing Systems},
  volume={37},
  pages={24081--24125},
  year={2024}
}

@article{song2025history,
  title={History-guided video diffusion},
  author={Song, Kiwhan and Chen, Boyuan and Simchowitz, Max and Du, Yilun and Tedrake, Russ and Sitzmann, Vincent},
  journal={arXiv preprint arXiv:2502.06764},
  year={2025}
}

@article{wu2025pack,
  title={Pack and force your memory: Long-form and consistent video generation},
  author={Wu, Xiaofei and Zhang, Guozhen and Xu, Zhiyong and Zhou, Yuan and Lu, Qinglin and He, Xuming},
  journal={arXiv preprint arXiv:2510.01784},
  year={2025}
}

@article{wang2026worldcompass,
  title={WorldCompass: Reinforcement Learning for Long-Horizon World Models},
  author={Wang, Zehan and Wang, Tengfei and Zhang, Haiyu and Zuo, Xuhui and Wu, Junta and Wang, Haoyuan and Sun, Wenqiang and Wang, Zhenwei and Cao, Chenjie and Zhao, Hengshuang and others},
  journal={arXiv preprint arXiv:2602.09022},
  year={2026}
}

@article{xiang2025pan,
  title={Pan: A world model for general, interactable, and long-horizon world simulation},
  author={Xiang, Jiannan and Gu, Yi and Liu, Zihan and Feng, Zeyu and Gao, Qiyue and Hu, Yiyan and Huang, Benhao and Liu, Guangyi and Yang, Yichi and Zhou, Kun and others},
  journal={arXiv preprint arXiv:2511.09057},
  year={2025}
}

@article{team2026advancing,
  title={Advancing Open-source World Models},
  author={Team, Robbyant and Gao, Zelin and Wang, Qiuyu and Zeng, Yanhong and Zhu, Jiapeng and Cheng, Ka Leong and Li, Yixuan and Wang, Hanlin and Xu, Yinghao and Ma, Shuailei and others},
  journal={arXiv preprint arXiv:2601.20540},
  year={2026}
}

@article{samuel2026fast,
  title={Fast Autoregressive Video Diffusion and World Models with Temporal Cache Compression and Sparse Attention},
  author={Samuel, Dvir and Tzachor, Issar and Levy, Matan and Green, Micahel and Chechik, Gal and Ben-Ari, Rami},
  journal={arXiv preprint arXiv:2602.01801},
  year={2026}
}

@article{chen2026past,
  title={Past-and Future-Informed KV Cache Policy with Salience Estimation in Autoregressive Video Diffusion},
  author={Chen, Hanmo and Xu, Chenghao and Yang, Xu and Chen, Xuan and Deng, Cheng},
  journal={arXiv preprint arXiv:2601.21896},
  year={2026}
}

@article{ma2026flow,
  title={Flow caching for autoregressive video generation},
  author={Ma, Yuexiao and Zheng, Xuzhe and Xu, Jing and Xu, Xiwei and Ling, Feng and Zheng, Xiawu and Kuang, Huafeng and Li, Huixia and Wang, Xing and Xiao, Xuefeng and others},
  journal={arXiv preprint arXiv:2602.10825},
  year={2026}
}

@article{huang2025spacetimepilot,
  title={SpaceTimePilot: Generative Rendering of Dynamic Scenes Across Space and Time},
  author={Huang, Zhening and Jeong, Hyeonho and Chen, Xuelin and Gryaditskaya, Yulia and Wang, Tuanfeng Y and Lasenby, Joan and Huang, Chun-Hao},
  journal={arXiv preprint arXiv:2512.25075},
  year={2025}
}

@inproceedings{liang2025wonderland,
  title={Wonderland: Navigating 3d scenes from a single image},
  author={Liang, Hanwen and Cao, Junli and Goel, Vidit and Qian, Guocheng and Korolev, Sergei and Terzopoulos, Demetri and Plataniotis, Konstantinos N and Tulyakov, Sergey and Ren, Jian},
  booktitle={Proceedings of the Computer Vision and Pattern Recognition Conference},
  pages={798--810},
  year={2025}
}

@inproceedings{yu2025wonderworld,
  title={Wonderworld: Interactive 3d scene generation from a single image},
  author={Yu, Hong-Xing and Duan, Haoyi and Herrmann, Charles and Freeman, William T and Wu, Jiajun},
  booktitle={Proceedings of the Computer Vision and Pattern Recognition Conference},
  pages={5916--5926},
  year={2025}
}

@InProceedings{li2025wonderplay,
    title     = {WonderPlay: Dynamic 3D Scene Generation from a Single Image and Actions},
    author    = {Li, Zizhang and Yu, Hong-Xing and Liu, Wei and Yang, Yin and Herrmann, Charles and Wetzstein, Gordon and Wu, Jiajun},
    booktitle = {Proceedings of the IEEE/CVF international conference on computer vision},
    year      = {2025},
}

@article{hu2025ex,
  title={Ex-4d: Extreme viewpoint 4d video synthesis via depth watertight mesh},
  author={Hu, Tao and Peng, Haoyang and Liu, Xiao and Ma, Yuewen},
  journal={arXiv preprint arXiv:2506.05554},
  year={2025}
}

@article{liao2025thinking,
  title={Thinking with Camera: A Unified Multimodal Model for Camera-Centric Understanding and Generation},
  author={Liao, Kang and Wu, Size and Wu, Zhonghua and Jin, Linyi and Wang, Chao and Wang, Yikai and Wang, Fei and Li, Wei and Loy, Chen Change},
  journal={arXiv preprint arXiv:2510.08673},
  year={2025}
}

@article{song2025worldforge,
  title={Worldforge: Unlocking emergent 3d/4d generation in video diffusion model via training-free guidance},
  author={Song, Chenxi and Yang, Yanming and Zhao, Tong and Li, Ruibo and Zhang, Chi},
  journal={arXiv preprint arXiv:2509.15130},
  year={2025}
}

@article{yang2026neoverse,
  title={NeoVerse: Enhancing 4D World Model with in-the-wild Monocular Videos},
  author={Yang, Yuxue and Fan, Lue and Shi, Ziqi and Peng, Junran and Wang, Feng and Zhang, Zhaoxiang},
  journal={arXiv preprint arXiv:2601.00393},
  year={2026}
}

@inproceedings{burgert2025go,
  title={Go-with-the-flow: Motion-controllable video diffusion models using real-time warped noise},
  author={Burgert, Ryan and Xu, Yuancheng and Xian, Wenqi and Pilarski, Oliver and Clausen, Pascal and He, Mingming and Ma, Li and Deng, Yitong and Li, Lingxiao and Mousavi, Mohsen and others},
  booktitle={Proceedings of the Computer Vision and Pattern Recognition Conference},
  pages={13--23},
  year={2025}
}

@article{huang2025voyager,
  title={Voyager: Long-range and world-consistent video diffusion for explorable 3d scene generation},
  author={Huang, Tianyu and Zheng, Wangguandong and Wang, Tengfei and Liu, Yuhao and Wang, Zhenwei and Wu, Junta and Jiang, Jie and Li, Hui and Lau, Rynson and Zuo, Wangmeng and others},
  journal={ACM Transactions on Graphics (TOG)},
  volume={44},
  number={6},
  pages={1--15},
  year={2025},
  publisher={ACM New York, NY, USA}
}

@inproceedings{luo2025camclonemaster,
  title={Camclonemaster: Enabling reference-based camera control for video generation},
  author={Luo, Yawen and Shi, Xiaoyu and Bai, Jianhong and Xia, Menghan and Xue, Tianfan and Wang, Xintao and Wan, Pengfei and Zhang, Di and Gai, Kun},
  booktitle={Proceedings of the SIGGRAPH Asia 2025 Conference Papers},
  pages={1--10},
  year={2025}
}

@article{wu2025geometry,
  title={Geometry forcing: Marrying video diffusion and 3d representation for consistent world modeling},
  author={Wu, Haoyu and Wu, Diankun and He, Tianyu and Guo, Junliang and Ye, Yang and Duan, Yueqi and Bian, Jiang},
  journal={arXiv preprint arXiv:2507.07982},
  year={2025}
}

@article{xiao2025video4spatial,
  title={Video4Spatial: Towards Visuospatial Intelligence with Context-Guided Video Generation},
  author={Xiao, Zeqi and Zhao, Yiwei and Li, Lingxiao and Lan, Yushi and Yu, Ning and Garg, Rahul and Cooper, Roshni and Taghavi, Mohammad H and Pan, Xingang},
  journal={arXiv preprint arXiv:2512.03040},
  year={2025}
}

@article{fortier2025gimbaldiffusion,
  title={GimbalDiffusion: Gravity-Aware Camera Control for Video Generation},
  author={Fortier-Chouinard, Fr{\'e}d{\'e}ric and Hold-Geoffroy, Yannick and Deschaintre, Valentin and Gadelha, Matheus and Lalonde, Jean-Fran{\c{c}}ois},
  journal={arXiv preprint arXiv:2512.09112},
  year={2025}
}

@article{lee20253d,
  title={3D Scene Prompting for Scene-Consistent Camera-Controllable Video Generation},
  author={Lee, JoungBin and Jung, Jaewoo and Han, Jisang and Narihira, Takuya and Fukuda, Kazumi and Seo, Junyoung and Hong, Sunghwan and Mitsufuji, Yuki and Kim, Seungryong},
  journal={arXiv preprint arXiv:2510.14945},
  year={2025}
}

@inproceedings{zhang2025world,
  title={World-consistent video diffusion with explicit 3d modeling},
  author={Zhang, Qihang and Zhai, Shuangfei and Martin, Miguel Angel Bautista and Miao, Kevin and Toshev, Alexander and Susskind, Joshua and Gu, Jiatao},
  booktitle={Proceedings of the Computer Vision and Pattern Recognition Conference},
  pages={21685--21695},
  year={2025}
}

@article{li2025recamdriving,
  title={ReCamDriving: LiDAR-Free Camera-Controlled Novel Trajectory Video Generation},
  author={Li, Yaokun and Wang, Shuaixian and Guo, Mantang and Huang, Jiehui and Ding, Taojun and Hu, Mu and Wang, Kaixuan and Shen, Shaojie and Tan, Guang},
  journal={arXiv preprint arXiv:2512.03621},
  year={2025}
}

@article{van2026anyview,
  title={AnyView: Synthesizing Any Novel View in Dynamic Scenes},
  author={Van Hoorick, Basile and Chen, Dian and Iwase, Shun and Tokmakov, Pavel and Irshad, Muhammad Zubair and Vasiljevic, Igor and Gupta, Swati and Cheng, Fangzhou and Zakharov, Sergey and Guizilini, Vitor Campagnolo},
  journal={arXiv preprint arXiv:2601.16982},
  year={2026}
}

@article{lu2025see4d,
  title={SEE4D: Pose-Free 4D Generation via Auto-Regressive Video Inpainting},
  author={Lu, Dongyue and Liang, Ao and Huang, Tianxin and Fu, Xiao and Zhao, Yuyang and Ma, Baorui and Pan, Liang and Yin, Wei and Kong, Lingdong and Ooi, Wei Tsang and others},
  journal={arXiv preprint arXiv:2510.26796},
  year={2025}
}

@article{chou2025captain,
  title={Captain Safari: A World Engine},
  author={Chou, Yu-Cheng and Wang, Xingrui and Li, Yitong and Wang, Jiahao and Liu, Hanting and Xie, Cihang and Yuille, Alan and Xiao, Junfei},
  journal={arXiv preprint arXiv:2511.22815},
  year={2025}
}

@article{xie2026lavr,
  title={LaVR: Scene Latent Conditioned Generative Video Trajectory Re-Rendering using Large 4D Reconstruction Models},
  author={Xie, Mingyang and Khan, Numair and Wang, Tianfu and Dhingra, Naina and Nam, Seonghyeon and Yang, Haitao and Hui, Zhuo and Metzler, Christopher and Vedaldi, Andrea and Pirsiavash, Hamed and others},
  journal={arXiv preprint arXiv:2601.14674},
  year={2026}
}

@article{wang2025taming,
  title={Taming Camera-Controlled Video Generation with Verifiable Geometry Reward},
  author={Wang, Zhaoqing and Xia, Xiaobo and Bie, Zhuolin and Liu, Jinlin and Yu, Dongdong and Bian, Jia-Wang and Wang, Changhu},
  journal={arXiv preprint arXiv:2512.02870},
  year={2025}
}

@article{park2025zero4d,
  title={Zero4D: Training-Free 4D Video Generation From Single Video Using Off-the-Shelf Video Diffusion},
  author={Park, Jangho and Kwon, Taesung and Ye, Jong Chul},
  journal={arXiv preprint arXiv:2503.22622},
  year={2025}
}

@article{chen2025postcam,
  title={PostCam: Camera-Controllable Novel-View Video Generation with Query-Shared Cross-Attention},
  author={Chen, Yipeng and Ye, Zhichao and Fang, Zhenzhou and Chen, Xinyu and Zhang, Xiaoyu and Liu, Jialing and Wang, Nan and Liu, Haomin and Zhang, Guofeng},
  journal={arXiv preprint arXiv:2511.17185},
  year={2025}
}

@article{pan2025diff4splat,
  title={Diff4Splat: Controllable 4D Scene Generation with Latent Dynamic Reconstruction Models},
  author={Pan, Panwang and Lin, Chenguo and Zhao, Jingjing and Li, Chenxin and Lin, Yuchen and Li, Haopeng and Yan, Honglei and Wen, Kairun and Lin, Yunlong and Yuan, Yixuan and others},
  journal={arXiv preprint arXiv:2511.00503},
  year={2025}
}

@article{zhang2025unified,
  title={Unified Camera Positional Encoding for Controlled Video Generation},
  author={Zhang, Cheng and Li, Boying and Wei, Meng and Cao, Yan-Pei and Gambardella, Camilo Cruz and Phung, Dinh and Cai, Jianfei},
  journal={arXiv preprint arXiv:2512.07237},
  year={2025}
}

@article{bian2025gs,
  title={GS-DiT: Advancing Video Generation with Pseudo 4D Gaussian Fields through Efficient Dense 3D Point Tracking},
  author={Bian, Weikang and Huang, Zhaoyang and Shi, Xiaoyu and Li, Yijin and Wang, Fu-Yun and Li, Hongsheng},
  journal={arXiv preprint arXiv:2501.02690},
  year={2025}
}

@String(CVPR= {IEEE Conf. Comput. Vis. Pattern Recog.})

@String(TOG= {ACM Trans. Graph.})

@String(ICLR = {Int. Conf. Learn. Represent.})

@inproceedings{LDM,
  title={High-resolution image synthesis with latent diffusion models},
  author={Rombach, Robin and Blattmann, Andreas and Lorenz, Dominik and Esser, Patrick and Ommer, Bj{\"o}rn},
  booktitle={Proceedings of the IEEE/CVF Conference on Computer Vision and Pattern Recognition},
  pages={10684--10695},
  year={2022}
}

@article{liu2024reconx,
  title={{ReconX:} Reconstruct Any Scene from Sparse Views with Video Diffusion Model},
  author={Liu, Fangfu and Sun, Wenqiang and Wang, Hanyang and Wang, Yikai and Sun, Haowen and Ye, Junliang and Zhang, Jun and Duan, Yueqi},
  journal={arXiv preprint arXiv:2408.16767},
  year={2024}
}

@inproceedings{huang2024vbench,
  title={Vbench: Comprehensive benchmark suite for video generative models},
  author={Huang, Ziqi and He, Yinan and Yu, Jiashuo and Zhang, Fan and Si, Chenyang and Jiang, Yuming and Zhang, Yuanhan and Wu, Tianxing and Jin, Qingyang and Chanpaisit, Nattapol and others},
  booktitle={Proceedings of the IEEE/CVF Conference on Computer Vision and Pattern Recognition},
  pages={21807--21818},
  year={2024}
}

@inproceedings{deitke2023objaverse,
  title={Objaverse: A universe of annotated 3d objects},
  author={Deitke, Matt and Schwenk, Dustin and Salvador, Jordi and Weihs, Luca and Michel, Oscar and VanderBilt, Eli and Schmidt, Ludwig and Ehsani, Kiana and Kembhavi, Aniruddha and Farhadi, Ali},
  booktitle={Proceedings of the IEEE/CVF conference on computer vision and pattern recognition},
  pages={13142--13153},
  year={2023}
}

@article{yu20244real,
  title={4real: Towards photorealistic 4d scene generation via video diffusion models},
  author={Yu, Heng and Wang, Chaoyang and Zhuang, Peiye and Menapace, Willi and Siarohin, Aliaksandr and Cao, Junli and Jeni, L{\'a}szl{\'o} and Tulyakov, Sergey and Lee, Hsin-Ying},
  journal={Advances in Neural Information Processing Systems},
  volume={37},
  pages={45256--45280},
  year={2024}
}

@article{yang2025omnicam,
  title={OmniCam: Unified Multimodal Video Generation via Camera Control},
  author={Yang, Xiaoda and Xu, Jiayang and Luan, Kaixuan and Zhan, Xinyu and Qiu, Hongshun and Shi, Shijun and Li, Hao and Yang, Shuai and Zhang, Li and Yu, Checheng and others},
  journal={arXiv preprint arXiv:2504.02312},
  year={2025}
}

@article{bernal2025precisecam,
  title={PreciseCam: Precise Camera Control for Text-to-Image Generation},
  author={Bernal-Berdun, Edurne and Serrano, Ana and Masia, Belen and Gadelha, Matheus and Hold-Geoffroy, Yannick and Sun, Xin and Gutierrez, Diego},
  journal={arXiv preprint arXiv:2501.12910},
  year={2025}
}

@article{wang20244real,
  title={4Real-Video: Learning Generalizable Photo-Realistic 4D Video Diffusion},
  author={Wang, Chaoyang and Zhuang, Peiye and Ngo, Tuan Duc and Menapace, Willi and Siarohin, Aliaksandr and Vasilkovsky, Michael and Skorokhodov, Ivan and Tulyakov, Sergey and Wonka, Peter and Lee, Hsin-Ying},
  journal={arXiv preprint arXiv:2412.04462},
  year={2024}
}

@article{xu2024camco,
  title={CamCo: Camera-Controllable 3D-Consistent Image-to-Video Generation},
  author={Xu, Dejia and Nie, Weili and Liu, Chao and Liu, Sifei and Kautz, Jan and Wang, Zhangyang and Vahdat, Arash},
  journal={arXiv preprint arXiv:2406.02509},
  year={2024}
}

@inproceedings{muller2024multidiff,
  title={MultiDiff: Consistent Novel View Synthesis from a Single Image},
  author={M{\"u}ller, Norman and Schwarz, Katja and R{\"o}ssle, Barbara and Porzi, Lorenzo and Bul{\`o}, Samuel Rota and Nie{\ss}ner, Matthias and Kontschieder, Peter},
  booktitle={Proc. CVPR},
  year={2024}
}

@inproceedings{gao2024cat3d,
  title={Cat3d: Create anything in 3d with multi-view diffusion models},
  author={Gao, Ruiqi and Holynski, Aleksander and Henzler, Philipp and Brussee, Arthur and Martin-Brualla, Ricardo and Srinivasan, Pratul and Barron, Jonathan T and Poole, Ben},
  booktitle={Proc. NeurIPS},
  year={2024}
}

@article{zhang2024recapture,
  title={ReCapture: Generative Video Camera Controls for User-Provided Videos using Masked Video Fine-Tuning},
  author={Zhang, David Junhao and Paiss, Roni and Zada, Shiran and Karnad, Nikhil and Jacobs, David E and Pritch, Yael and Mosseri, Inbar and Shou, Mike Zheng and Wadhwa, Neal and Ruiz, Nataniel},
  journal={arXiv preprint arXiv:2411.05003},
  year={2024}
}

@article{wu2024cat4d,
  title={Cat4d: Create anything in 4d with multi-view video diffusion models},
  author={Wu, Rundi and Gao, Ruiqi and Poole, Ben and Trevithick, Alex and Zheng, Changxi and Barron, Jonathan T and Holynski, Aleksander},
  journal={Proc. CVPR},
  year={2025}
}

@article{bai2024syncammaster,
  title={SynCamMaster: Synchronizing Multi-Camera Video Generation from Diverse Viewpoints},
  author={Bai, Jianhong and Xia, Menghan and Wang, Xintao and Yuan, Ziyang and Fu, Xiao and Liu, Zuozhu and Hu, Haoji and Wan, Pengfei and Zhang, Di},
  journal={Proc. ICLR},
  year={2025}
}

@article{duan2026liveworld,
  title={LiveWorld: Simulating Out-of-Sight Dynamics in Generative Video World Models},
  author={Duan, Zicheng and Xia, Jiatong and Zhang, Zeyu and Zhang, Wenbo and Zhou, Gengze and Gou, Chenhui and He, Yefei and Chen, Feng and Zhang, Xinyu and Liu, Lingqiao},
  journal={arXiv preprint arXiv:2603.07145},
  year={2026}
}
}

\newpage
\appendix
\onecolumn

\section*{Appendix}

\section{Additional Implementation Details}
\label{additional_implementation}

\subsection{Training/Inference Details}

We train AnchorWeave on a subset of 10K videos sampled from RealEstate10K~\cite{zhou2018stereo} and DL3DV~\cite{ling2024dl3dv}. For each video, we estimate per-frame local geometry using TTT3R~\cite{chen2025ttt3r} and store the resulting point clouds as local spatial memories.

During training, we use a chunk length of $C=8$, retrieving memories once every 8 frames, and retrieve at most $K=4$ local point clouds for conditioning. To improve robustness, we randomly sample from the candidate local memory pool as a form of memory augmentation. We also apply random frame masking to anchor videos to enhance robustness to missing or imperfect geometric guidance. For memory retrieval, the first rendered frame is always selected as the initial anchor, following the standard anchor-video convention in which the initial frame serves as a stable geometric reference for subsequent generation.

We apply AnchorWeave to two DiT-based backbones: CogVideoX-I2V-5B~\cite{yang2024cogvideox} and Wan2.2-TI2V-5B~\cite{wan2025}. For Wan2.2, we train on 81-frame videos, while for CogVideoX we use 49-frame videos. During training, the backbone weights are frozen and only the newly introduced modules are optimized.

The weaving controller is injected into the first third of backbone layers to balance capacity and efficiency, and is applied during the first 80\% of denoising steps. The backbone latent receives two control signals: (1) fused anchor latents and (2) target camera pose embeddings. Both control signals use a control weight of 1.0 (i.e., added with equal weight to the original backbone features). 
Classifier-free guidance (CFG) follows the default configuration of each backbone.

We train using the Adam optimizer and the standard MSE loss with a learning rate of $2\times10^{-4}$ for 10K steps and a batch size of 8. Training is conducted on 8×H100 GPUs and takes approximately one day.

\subsection{Baseline Re-implementation}
\label{baselines}
\noindent\textbf{Context-as-Memory.}
We reimplement Context-as-Memory with CogVideoX backbone for fair comparison. We use the same training data as AnchorWeave, sampled from RealEstate10K and DL3DV. For each video, we follow Context-as-Memory to select context frames with high trajectory overlap with the target segment. Specifically, we sample 12 context frames per video as conditioning inputs. We apply LoRA to fine-tune the entire backbone for multi-context conditioning, with LoRA rank 16 and $\alpha=32$.
Training is conducted for 10K steps with a batch size of 8, consistent with AnchorWeave. During inference, we select context frames with the highest overall trajectory overlap as conditioning inputs.

\noindent\textbf{SPMem.}
We reimplement SPMem using CogVideoX as the backbone and train on the same dataset as AnchorWeave (RealEstate10K and DL3DV).
SPMem conditions generation on both (1) rendered global point clouds via a ControlNet branch and (2) keyframe latents via additional cross-attention layers. We follow a similar structure: the ControlNet branch is injected into the first third of backbone layers, and an additional cross-attention layer is inserted into each backbone block to condition on keyframe latents.
Keyframes are selected using the same trajectory-overlap strategy as in Context-as-Memory. Global point clouds are constructed using TTT3R and rendered for conditioning.
During training, we freeze the backbone and only fine-tune the newly introduced cross-attention layers and the ControlNet branch. The model is trained for 10K steps with a batch size of 8, consistent with AnchorWeave.

\section{Coverage-Driven Memory Retrieval Details}
\label{retrieval details}

We demonstrate our memory retrieval method in the main paper.
Here we provide the detailed retrieval procedure in ~\Cref{alg:greedy_chunk_retrieval}.
In practice, the first rendered frame is always selected as the initial anchor, following the standard anchor-video convention where the initial frame serves as a stable geometric reference.
Coverage is computed on a compressed point cloud representation (16×–32× downsampled), making the retrieval process efficient in practice. 
For a typical video sequence, retrieval and rendering takes only a few seconds and does not constitute a computational bottleneck.

\begin{algorithm}
\caption{Coverage-driven 3D-aware memory retrieval for anchor-video construction}
\label{alg:greedy_chunk_retrieval}
\small
\begin{algorithmic}

\State \textbf{Input:} Local point clouds $\{\mathcal{P}_i\}$ with poses $\{T_i\}$; target trajectory $\{\hat{T}_t\}_{t=1}^{M}$; chunk length $D$; max anchors per chunk $K$
\State \textbf{Output:} Anchor videos and retrieved-to-target relative poses

\Statex
\State Partition target trajectory into ordered chunks $\{\mathcal{C}_m\}$, each with $D$ target poses

\For{each chunk $\mathcal{C}_m$}
    \State Initialize anchor set $\mathcal{A}_m \leftarrow \{\mathcal{P}_{\text{latest}}\}$

    \Statex
    \Statex \textbf{Candidate filtering:}
    \State Retain $\mathcal{P}_i$ whose pose $T_i$ has sufficient FOV overlap with chunk $\mathcal{C}_m$

    \While{$|\mathcal{A}_m| < K$ and coverage improves}
        \State Select $\mathcal{P}_i$ that maximizes additional visible coverage when rendered to $\mathcal{C}_m$
        \State $\mathcal{A}_m \leftarrow \mathcal{A}_m \cup \{\mathcal{P}_i\}$
    \EndWhile

    \State Compute retrieved-to-target relative poses for anchors in $\mathcal{A}_m$
\EndFor

\State Concatenate rendered anchors across chunks to form anchor videos and corresponding retrieved-to-target relative poses

\end{algorithmic}
\end{algorithm}

\section{Camera Controllability Comparison with Baselines}
\label{camera control}

\begin{table}[ht]
\centering
\caption{Camera control ability comparisons.
Lower is better.}
\label{tab:camera_error}
\small
\setlength{\tabcolsep}{8pt}
\resizebox{0.85\linewidth}{!}{\begin{tabular}{lcc}
\toprule
\textbf{Method} & \textbf{Rotation Error ($\downarrow$)} & \textbf{Translation Error ($\downarrow$)} \\
\midrule

ViewCrafter~\cite{yu2024viewcrafter} & 1.34 & 3.21 \\
Gen3C~\cite{ren2025gen3c}           & 0.97 & 1.98 \\
TrajCrafter~\cite{yu2025trajectorycrafter} & 1.08 & 2.57 \\
EPiC~\cite{wang2025epic}            & 0.84 & 2.01 \\
SPMem~\cite{wu2025spmem} & 1.21 & 2.99 \\
TrajCrafter~\cite{yu2025trajectorycrafter} & 1.08 & 2.57 \\
SEVA~\cite{zhou2025stable}           & 0.78 & 1.96 \\

\midrule
AnchorWeave (Ours, no retrieval) & \textbf{0.61} & \textbf{1.72} \\

\bottomrule
\end{tabular}
}
\end{table}

We compared AnchorWeave's context-following ability and generation quality with prior methods in the main paper. Here
we further compare camera controllability with prior methods using rotation and translation errors, following the evaluation protocol of EPiC. We randomly sample 300 videos from both DL3DV and RealEstate10K and evaluate single-image-based camera control accuracy. For all methods, only the first frame is provided as input (Basic Image-to-Video Camera Control setting) without any other information. Thus, for our method, the conditioning signal is the anchor video rendered from the first frame, i.e., \textit{AnchorWeave (no retrieval)}.

As shown in Table~\ref{tab:camera_error}, our method achieves the lowest rotation and translation errors among all baselines. Notably, although our model is trained to follow multiple anchors, it still outperforms prior approaches when only a single anchor is provided at test time. This is because AnchorWeave combines anchor-video-based visual guidance with explicit camera pose control. When camera motion is large, the geometry rendered from the first frame quickly moves out of view, causing methods that rely solely on anchor videos (e.g., ViewCrafter, TrajCrafter, EPiC, and Gen3C, SPMem) to lose controllability. In contrast, our method can still follow the target camera trajectory using pose guidance, resulting in significantly more accurate camera motion.

\section{AnchorWeave Generation Process Visualization}

\label{visualization}

\begin{figure*}
    \centering

        \includegraphics[width=1.0\linewidth]{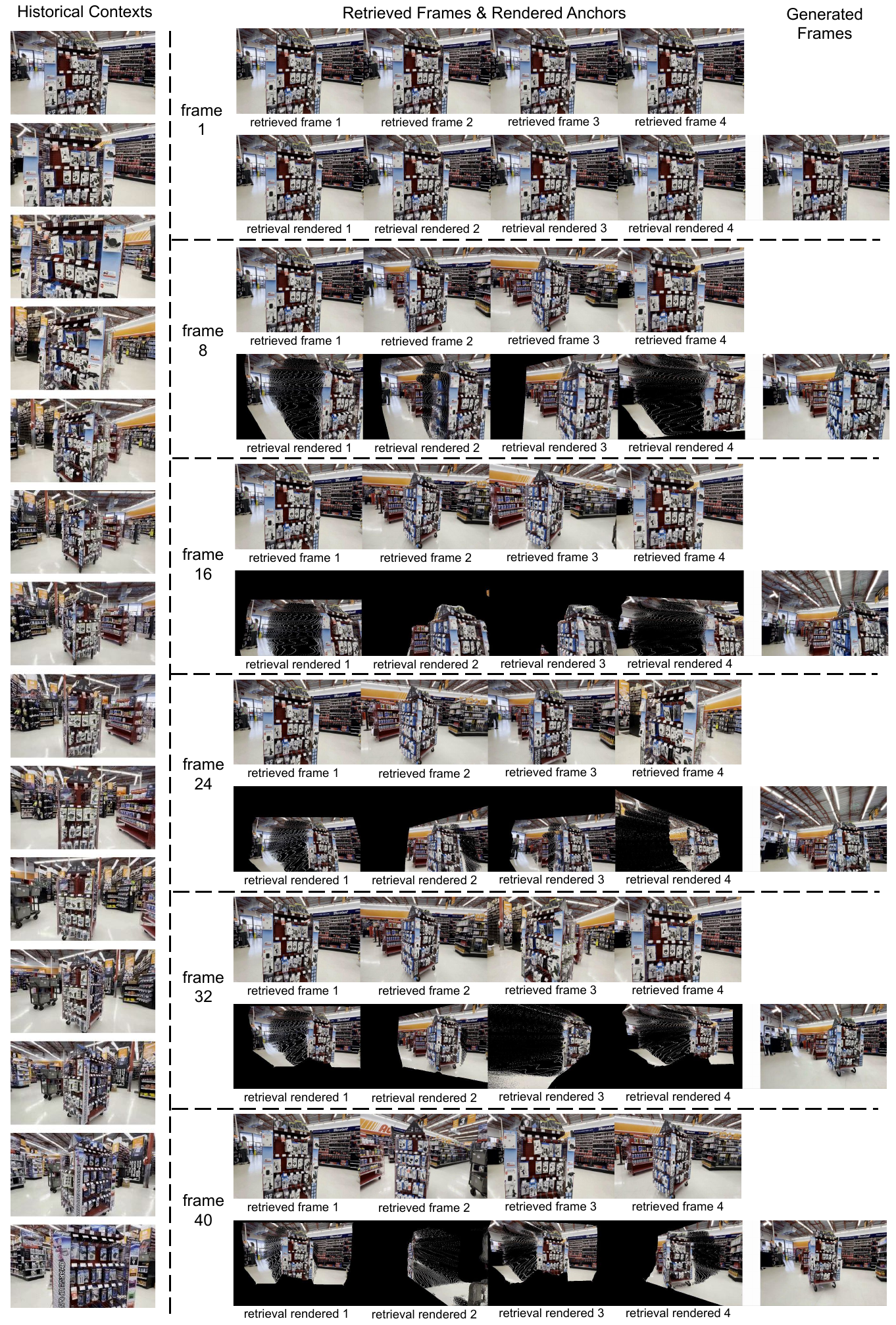}

    \caption{Context-conditioned camera control generation process of AnchorWeave.
    }
    \label{fig:weave_1}
\end{figure*}

\begin{figure*}
    \centering

        \includegraphics[width=1.0\linewidth]{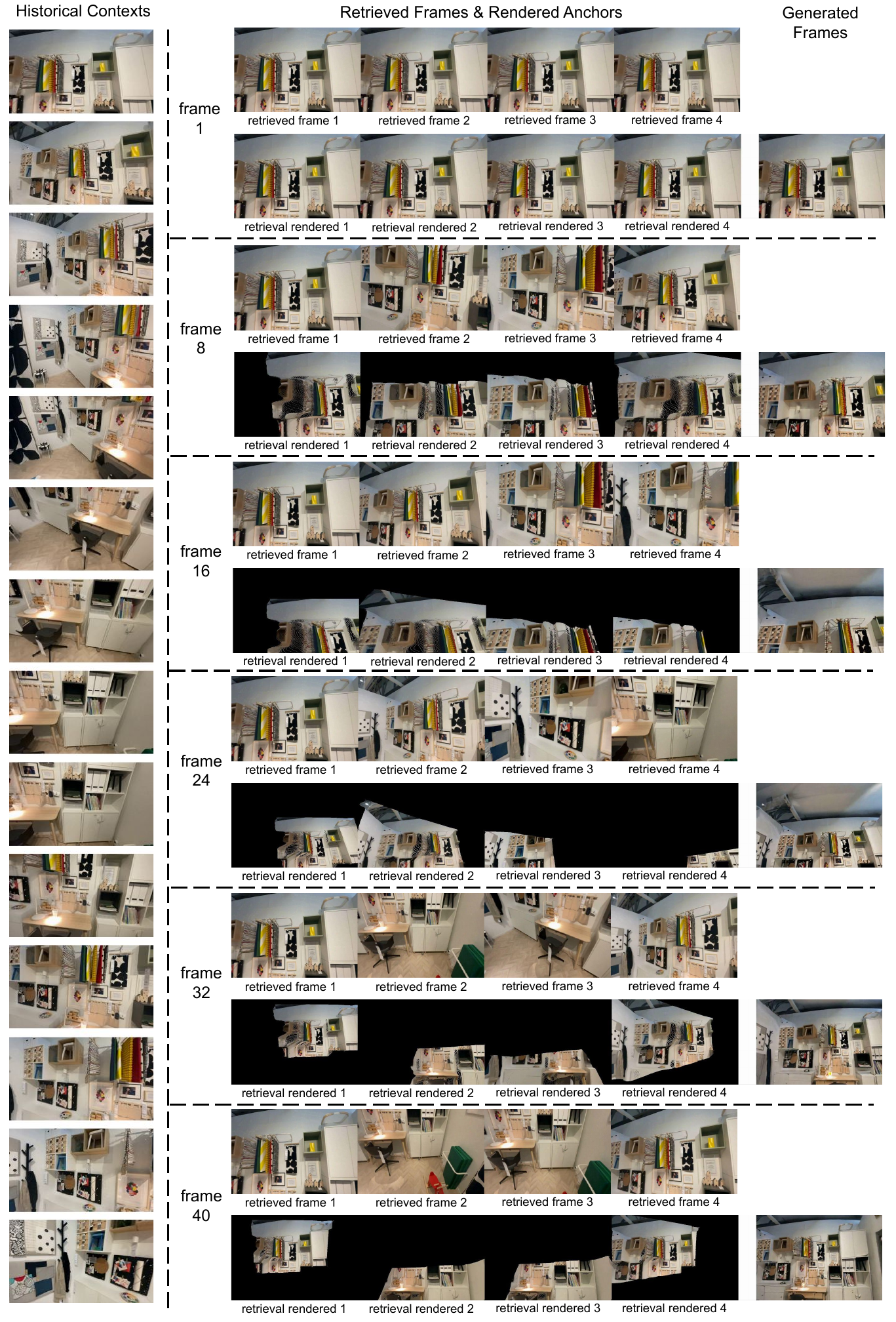}

    \caption{Context-conditioned camera control generation process of AnchorWeave.
    }
    \label{fig:weave_2}
\end{figure*}

We visualize the end-to-end AnchorWeave generation process in \Cref{fig:weave_1} and ~\Cref{fig:weave_2}.
We start from a historical context video sampled from DL3DV, and select one frame as the initial state.
The camera trajectory is then specified by a keyboard-style control sequence
(\texttt{move\_left} $\rightarrow$ \texttt{orient\_up} $\rightarrow$ \texttt{move\_backward} $\rightarrow$ \texttt{orient\_down} $\rightarrow$ \texttt{move\_right}, each for 8 steps),
which defines an interactive exploration of the environment.
To ensure long-horizon consistency with the history, AnchorWeave performs coverage-driven retrieval at each chunk:
it retrieves multiple history frames that jointly cover different visible regions under the target view, renders their corresponding geometric anchors, and weaves these anchors into the generation.
As shown in both examples, the retrieved frames cover complementary parts of the camera-visible area, and AnchorWeave can consistently fuse multiple anchors, producing frames that effectively merge information from different retrieved views.

\section{Qualitative Comparison with Baselines under Free-Form Camera Trajectory}
\label{free traj com}
\begin{figure*}
    \centering

        \includegraphics[width=0.99\linewidth]{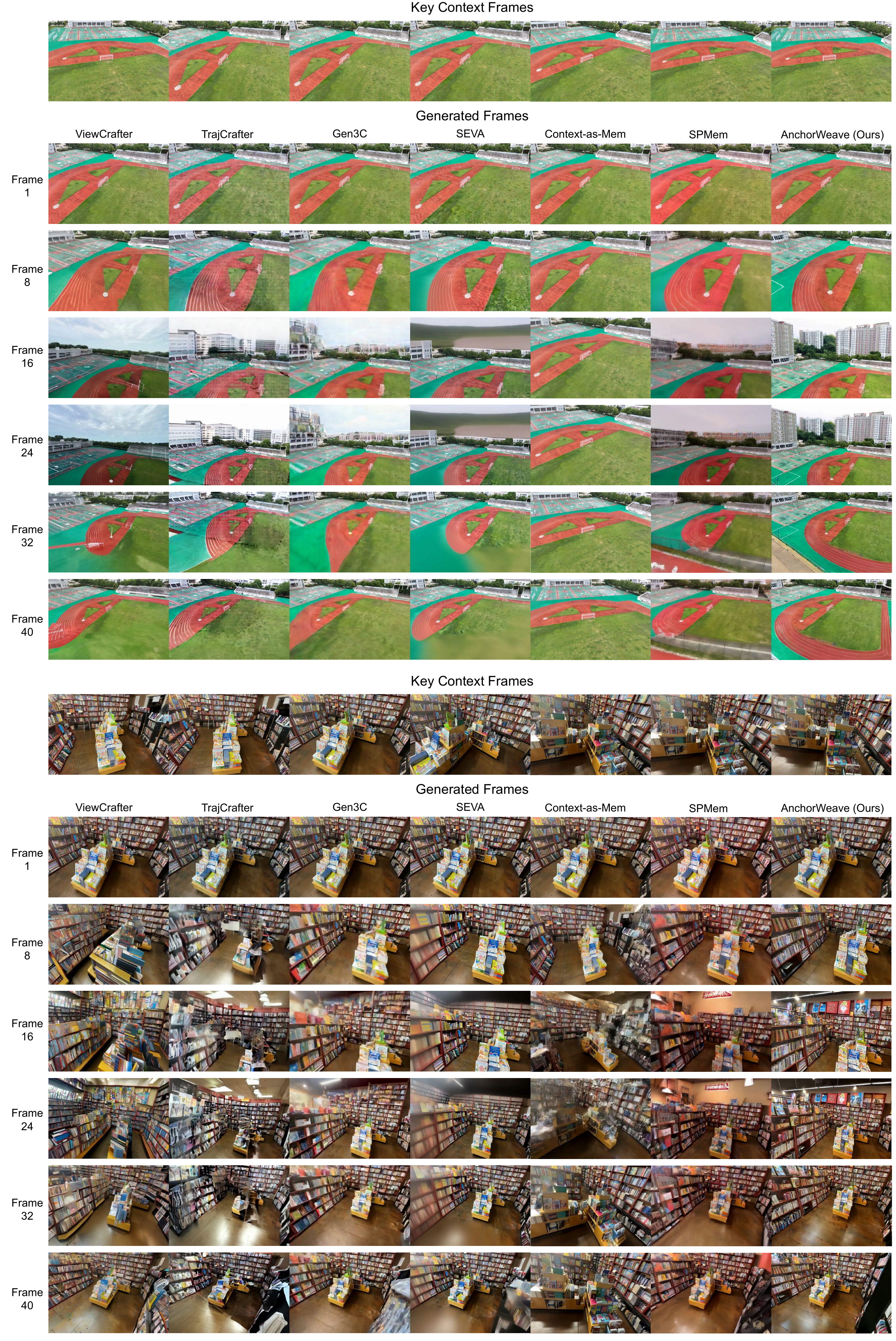}

    \caption{Context-conditioned camera control results of AnchorWeave compared with baselines.
    }
    \label{fig:keyboard_com_1}
\end{figure*}

\begin{figure*}
    \centering

        \includegraphics[width=1.0\linewidth]{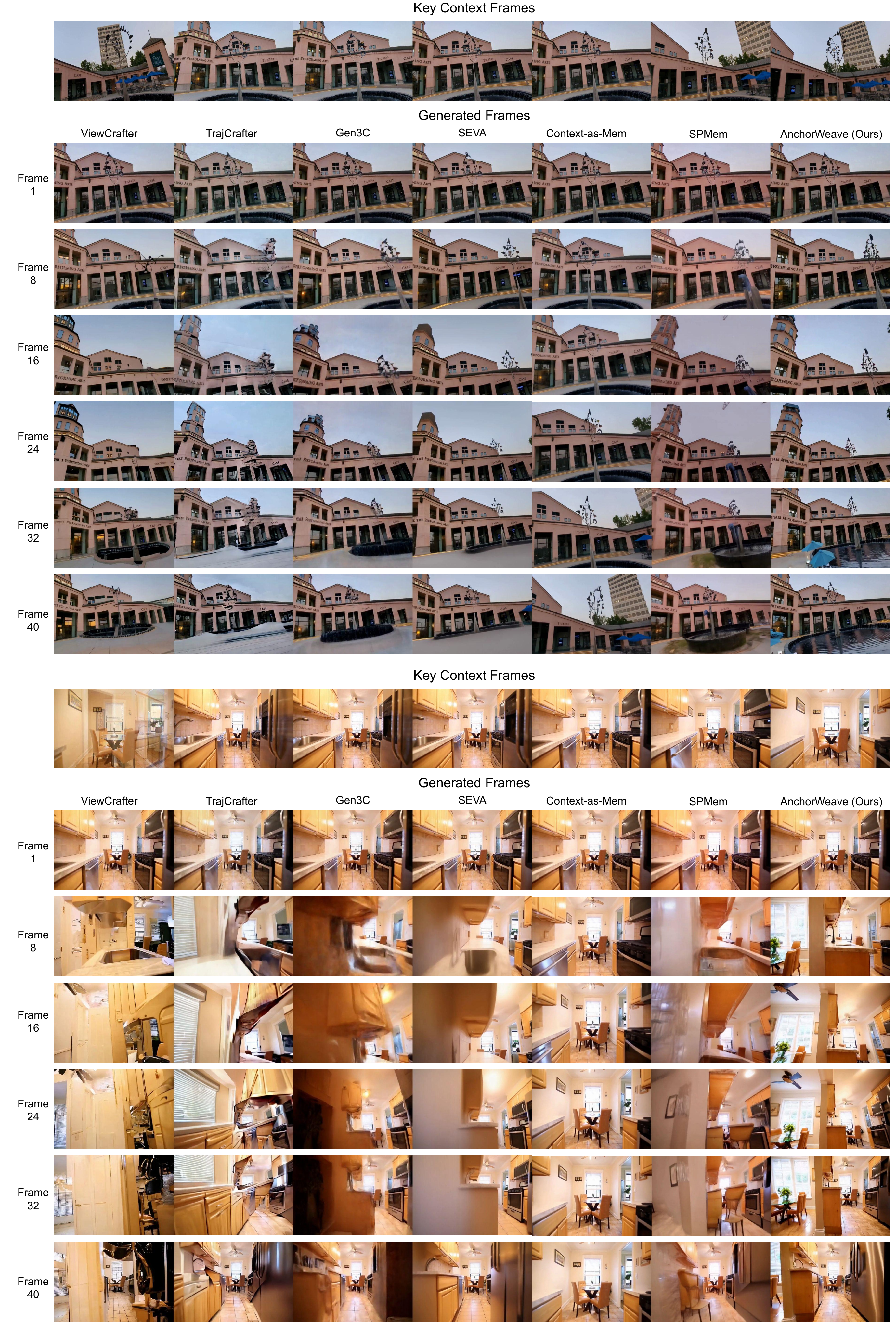}
        
    \caption{Context-conditioned camera control results of AnchorWeave compared with baselines.
    }
    \label{fig:keyboard_com_2}
\end{figure*}

\begin{figure*}
    \centering

        \includegraphics[width=1.0\linewidth]{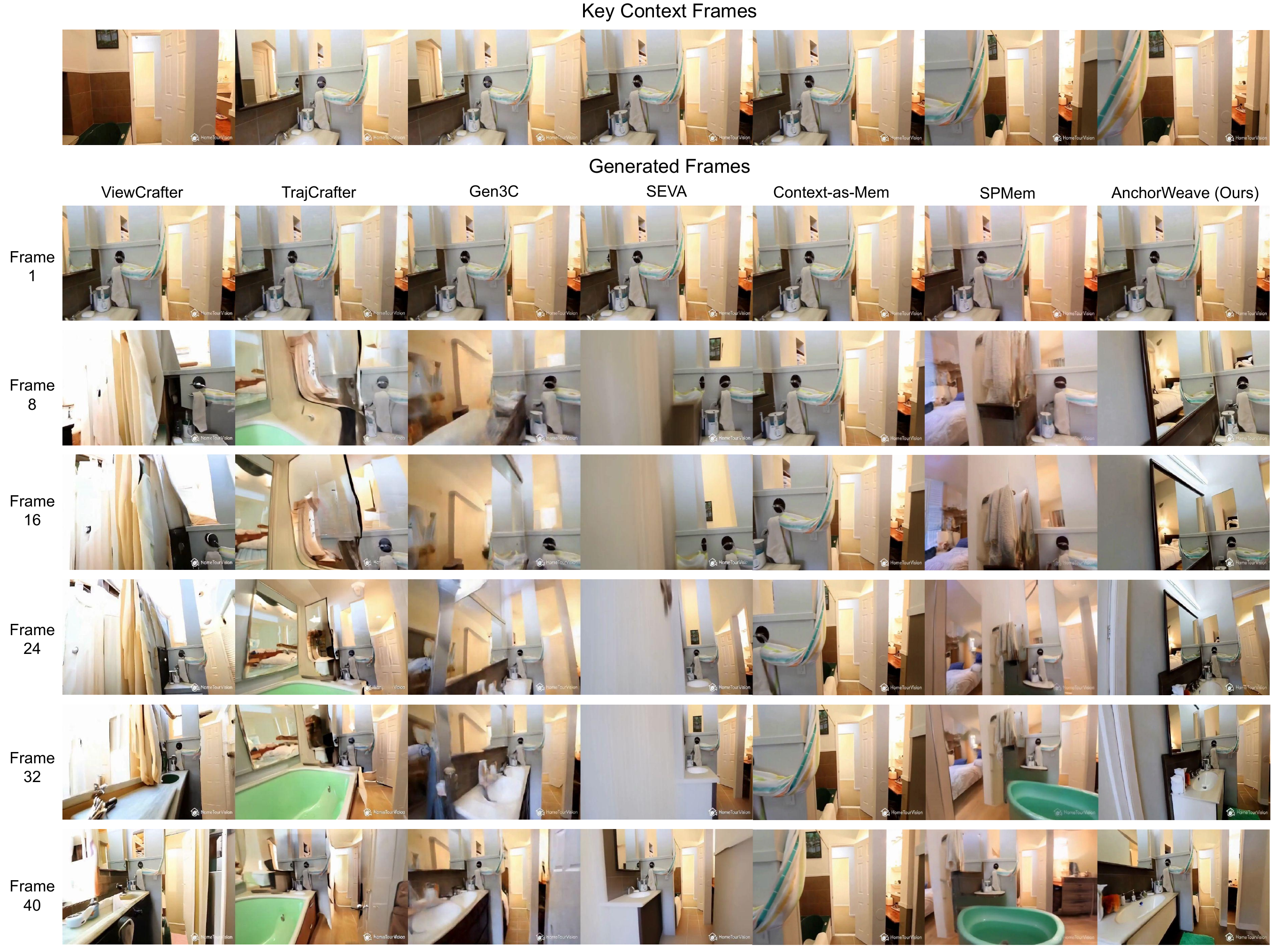}
        
    \caption{Context-conditioned camera control results of AnchorWeave compared with baselines.
    }
    \label{fig:keyboard_com_3}
\end{figure*}

We further compare AnchorWeave with ViewCrafter, TrajCrafter, Gen3C, SEVA, Context-as-Memory, and SPMem under a free-form, keyboard-controlled camera trajectory, as shown in \Cref{fig:keyboard_com_1,fig:keyboard_com_2,fig:keyboard_com_3}. 
Unlike the revisit setting where the camera replays the historical trajectory, here the motion follows a free-form control sequence (e.g., move\_left $\rightarrow$ orient\_up $\rightarrow$ move\_backward $\rightarrow$ orient\_down $\rightarrow$ move\_right), which deviates from the context trajectory and typical smooth motions observed in training data. 
We only show key historical context frames that spatially overlap with the commanded camera trajectory (Due to page limits, we sampled 7 frames per example).

ViewCrafter frequently introduces noticeable hallucinated structures when the camera moves into previously unseen regions.
TrajCrafter exhibits significant rendering artifacts and structural inconsistencies.
Gen3C tends to produce overly blurred outputs under large viewpoint changes.
SEVA maintains reasonable rendering quality in camera-visible regions but struggles to synthesize plausible content in newly exposed areas, leading to degraded details.
Context-as-Memory shows limited generalization to such out-of-distribution camera trajectories, often generating frames that implicitly follow the historical trajectory rather than the commanded motion.
SPMem suffers from severe hallucinations due to conditioning on imperfect global point cloud renderings.

In contrast, AnchorWeave remains stable under free camera control.
It faithfully preserves geometry in visible regions via anchor conditioning, while seamlessly synthesizing plausible content in newly exposed areas beyond the historical context.

\section{Long-Horizon Examples}

\label{long-horizon}

\begin{figure*}
    \centering

        \includegraphics[width=1.0\linewidth]{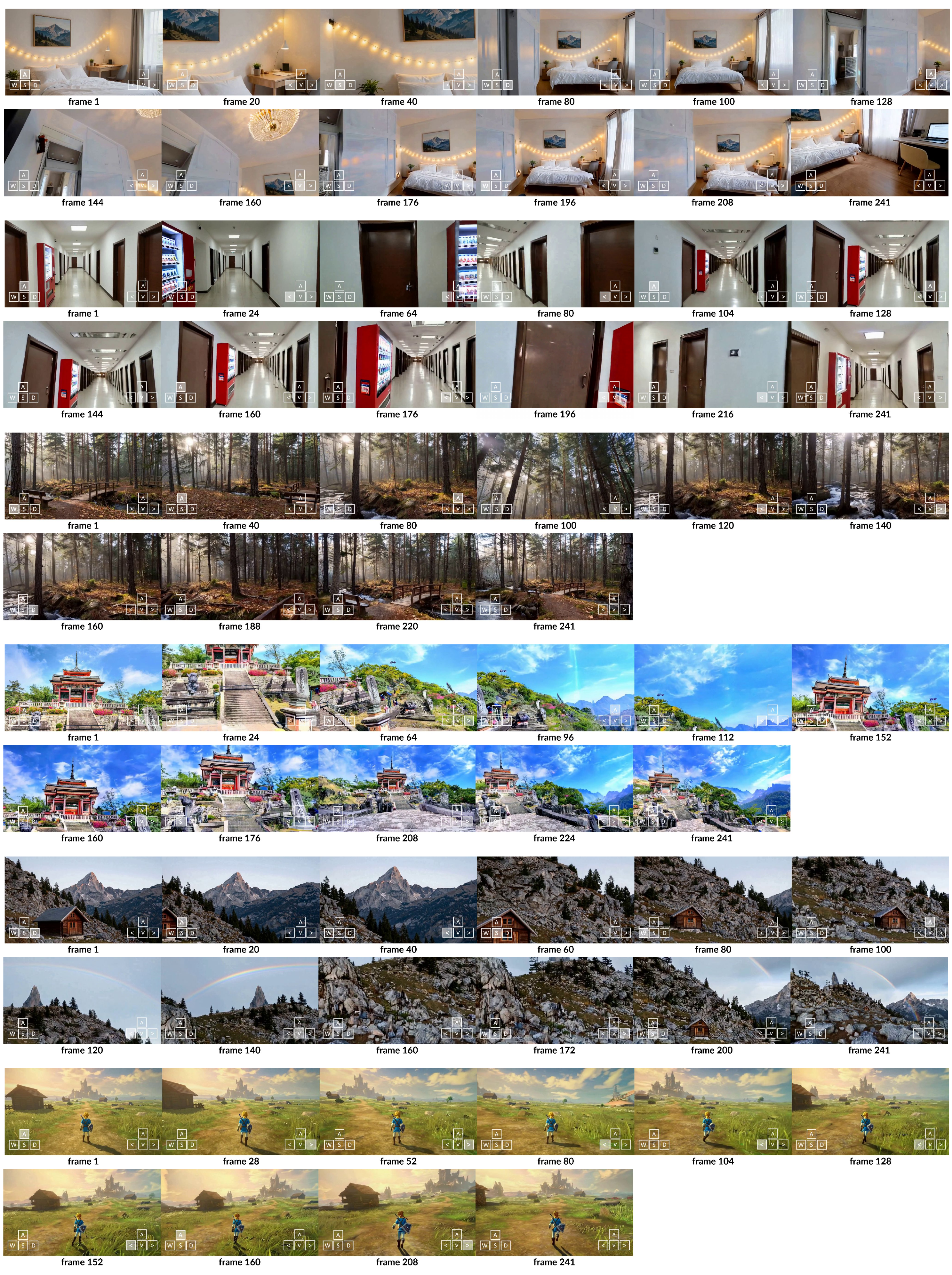}
        
    \caption{Long-horizon video generation on both open-domain static and dynamic environments.
    }
    \label{fig:long}
\end{figure*}

\Cref{fig:long} presents six additional long-horizon world exploration examples across diverse scenes, including both first-person and third-person settings. 
We define a fixed set of discrete camera control primitives, including rotational motions (up/down/left/right), translational motions (forward/backward/left/right), and a stop command. 
To account for scene scale differences (e.g., indoor vs.\ outdoor), the step magnitude is adjusted accordingly while keeping the action space unchanged. 
Each motion primitive can be executed for multiple consecutive steps. 
To maintain generation stability under continuous movement, we insert a brief stationary phase (four frames) after each motion segment, during which memory is updated.
We adopt Wan2.2 as the backbone and iteratively perform image-to-video generation with AnchorWeave for 81 frames conditioned on the last frame of the previous segment, resulting in 241 frames in total. 
For dynamic third-person scenes, we mask out the character to exclude it from the point-cloud memory construction to enable dynamic content generation.

\paragraph{Long-term consistency.}
Strong long-term consistency is observed across many examples. 
For instance, the table in the first scene, the chair and bridge in the third scene, the temple in the fourth scene, and the cabin in the fifth scene remain spatially consistent even after many frames and cross-segment viewpoint changes. 
When the camera revisits these regions, the objects still appear in coherent positions, demonstrating the effectiveness of AnchorWeave.

\paragraph{360° panoramic generation.}
The second example demonstrates full 360-degree rotation capability. 
Starting from a vending machine and corridor, the camera gradually rotates over many frames and eventually returns to the original viewpoint. 
The scene content remains consistent throughout the rotation, indicating robust global spatial coherence.

\paragraph{Third-person character control.}
The sixth example showcases a third-person game scenario. 
By masking the character and leveraging point-cloud control, we enable stable character motion aligned with the camera trajectory. 
Long-term consistency is also preserved: for example, after the character walks away in the first segment, the cabin and distant mountains remain visible when revisiting the area in later segments. 

Notably, this example is generated in a zero-shot manner. Our training data consist only of static-scene datasets (DL3DV and RealEstate10K), without any third-person character or dynamic gameplay supervision. The ability to generalize to character-centric, long-horizon scenarios therefore demonstrates the robustness of our design and its strong out-of-distribution generalization capability.

\section[Efficiency Scaling Details]{Efficiency Scaling Details}
\label{app:efficiency_scaling}

The main paper reports average retrieval and rendering cost for the full generation pipeline. Here, we further study how this cost scales with the amount of available history. For each history length, AnchorWeave retrieves a compact set of local memories using field-of-view overlap and renders the selected point clouds as conditioning anchors, whereas SPMem first merges all past observations into a global point cloud and renders from the accumulated scene representation. We measure the per-anchor retrieval-and-rendering time under the same generation setting used in the main paper.

\begin{figure}[H]
    \centering
    \includegraphics[width=0.55\linewidth]{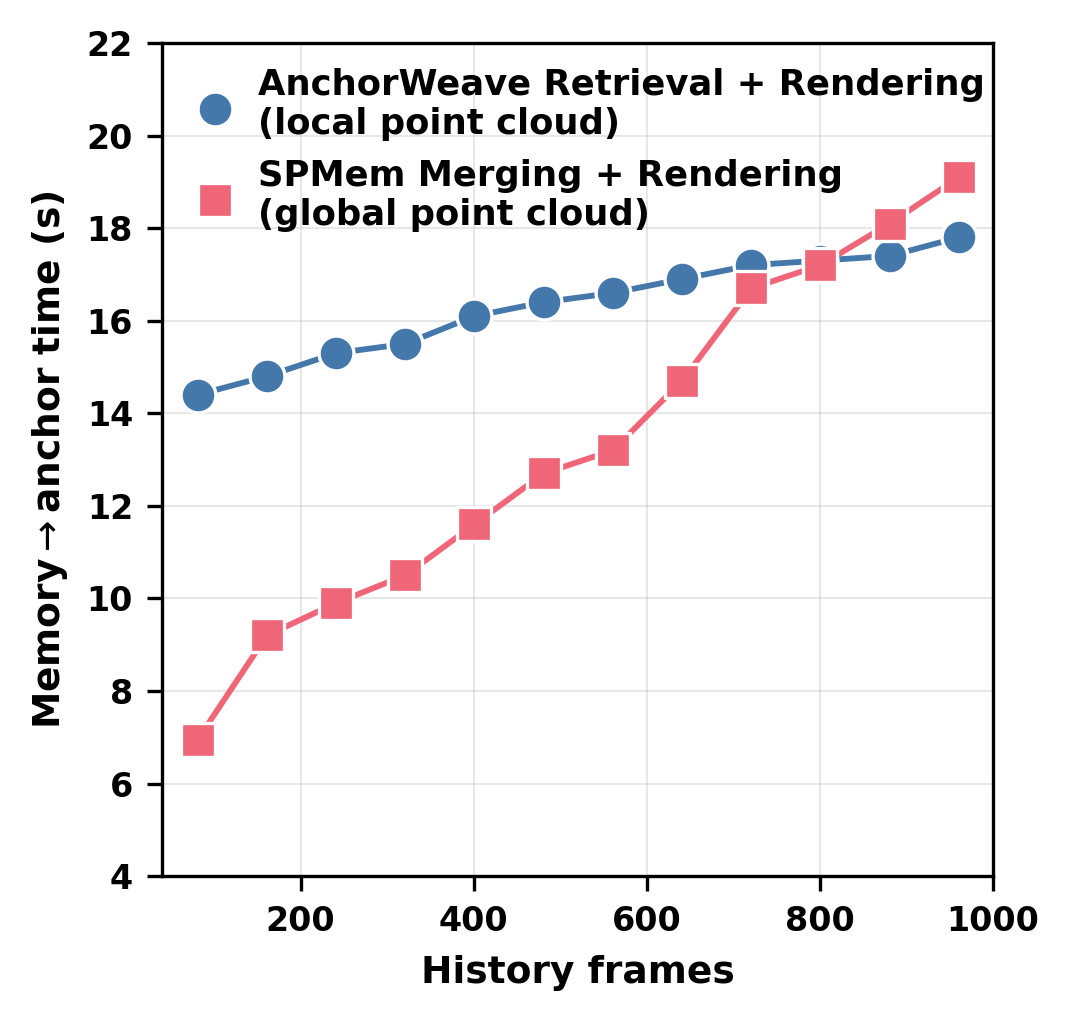}
    \caption{Runtime scaling of memory retrieval and rendering as history grows. AnchorWeave retrieves and renders from a compact set of local memories, while SPMem repeatedly merges and renders an accumulated global point cloud.}
    \label{fig:efficiency_scaling}
\end{figure}

\Cref{fig:efficiency_scaling} shows that AnchorWeave grows slowly as the history expands, since retrieval prunes irrelevant memories before rendering and coverage is computed on downsampled point clouds. In contrast, SPMem's cost increases more rapidly because the global point cloud becomes larger after each merged observation. This scaling behavior complements the average runtime comparison in the main paper: AnchorWeave adds moderate video-generation overhead for multi-anchor control, but keeps the memory retrieval stage stable over longer histories.

\FloatBarrier

\section[Additional Robustness Analysis]{Additional Robustness Analysis}
\label{app:robustness_analysis}

\subsection[Why Local Memories Outperform Global Fusion]{Why Local Memories Outperform Global Fusion}
We use a controlled rendering diagnostic to examine why local memories are more reliable than a single fused global memory. Starting from the same target view, we render conditioning images from point clouds constructed with different amounts of history. A point cloud built from one nearby frame is internally consistent and reaches 44.93 dB PSNR. When more frames are fused into a single global cloud, small per-frame errors in depth and pose accumulate as cross-view misalignment, reducing rendering quality to 16.15 dB after 45 frames. AnchorWeave avoids this degradation by keeping retrieved memories as separate view-consistent anchors, allowing the controller to combine complementary evidence without requiring a globally perfect fused reconstruction.

\begin{figure}[H]
    \centering
    \includegraphics[width=0.95\linewidth]{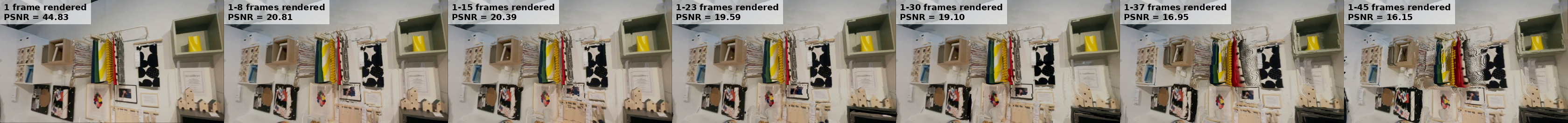}
    \caption{Diagnostic rendering quality as more frames are fused into a global point cloud. Global fusion accumulates alignment drift, while local memories retain the self-consistency of individual views.}
    \label{fig:global_drift_analysis}
\end{figure}

\subsection[Long-Horizon Stress Test]{Long-Horizon Stress Test}
The qualitative examples in \Cref{long-horizon} cover 241-frame trajectories. To probe longer histories, we construct an explore-and-revisit stress test from 20 initial images. Each cycle contains an 80-frame exploration phase, in which the camera moves and rotates until the initial view leaves the field of view, followed by an 80-frame revisit phase that returns to the original pose. Repeating this cycle produces sequences up to 2400 frames. Revisit consistency is evaluated by measuring PSNR between the generated revisit frame and the initial frame at each return point.

\begin{figure}[H]
    \centering
    \includegraphics[width=0.95\linewidth]{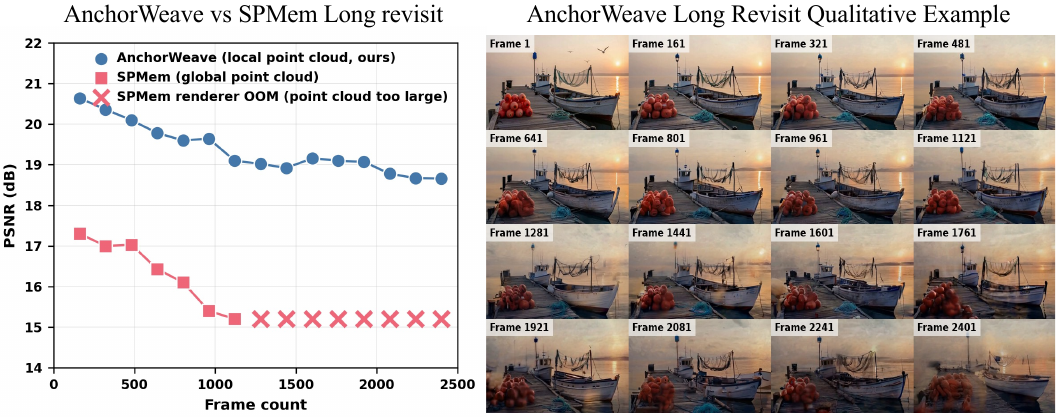}
    \caption{Long-horizon stress test with repeated explore-and-revisit loops. AnchorWeave maintains substantially better revisit consistency than SPMem over long histories.}
    \label{fig:long_stress}
\end{figure}

\Cref{fig:long_stress} shows that AnchorWeave maintains substantially higher revisit consistency than SPMem across long trajectories. SPMem degrades quickly as alignment errors accumulate in the global point cloud, and its rendering stage becomes impractical around frame 1120 because the merged point cloud is too large. AnchorWeave continues to operate by retrieving a compact subset of local memories and maintains roughly 20 dB revisit PSNR through the first 1K frames before gradually declining. The remaining degradation is mainly caused by synthesized errors being written back into memory and by long-horizon pose drift from the 3D estimator, which can make retrieved anchors increasingly difficult to reconcile.

\subsection[Dynamic Content]{Dynamic Content}
\begin{figure}[H]
    \centering
    \includegraphics[width=0.72\linewidth]{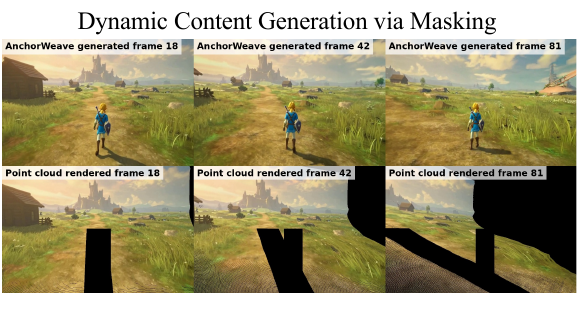}
    \caption{Dynamic-content handling via masking. Dynamic regions can be excluded before constructing the static point-cloud memory, allowing AnchorWeave to preserve a consistent geometric scaffold for the background.}
    \label{fig:dynamic_analysis}
\end{figure}

AnchorWeave is designed around static spatial memories, so independently moving foreground objects should not be written into the static point-cloud memory. For the third-person dynamic examples in \Cref{long-horizon}, we follow this principle by masking the character before memory construction, preserving the consistency of the background geometry while leaving character motion to the video generator. \Cref{fig:dynamic_analysis} illustrates this masking strategy.

\begin{figure}[H]
    \centering
    \includegraphics[width=0.95\linewidth]{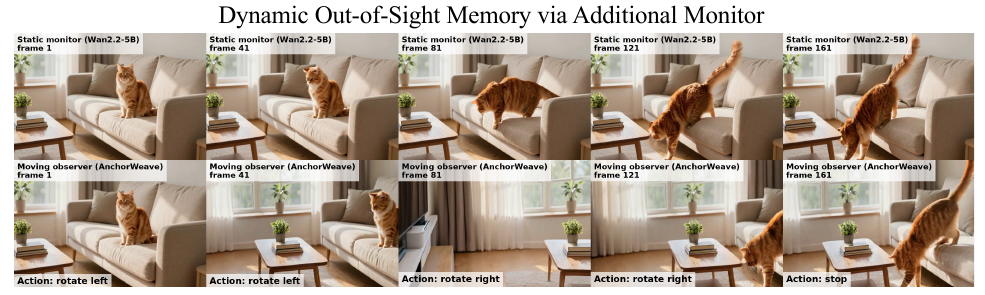}
    \caption{Dynamic out-of-sight memory via an additional monitor. A monitor-style dynamic-memory extension can update object state outside the current observer view, while AnchorWeave maintains the static spatial scaffold.}
    \label{fig:dynamic_monitor}
\end{figure}

More general dynamic-world modeling requires state updates for objects that move while outside the observer's view. This is complementary to AnchorWeave's static spatial scaffold: a monitor-style dynamic memory module, such as the one introduced by LiveWorld~\cite{duan2026liveworld}, could track dynamic entities and provide updated object states when the generated camera revisits them, while AnchorWeave continues to provide geometrically consistent static context. \Cref{fig:dynamic_monitor} sketches this extension.

\subsection[Failure Modes]{Failure Modes}

\begin{figure}[H]
    \centering
    \includegraphics[width=0.95\linewidth]{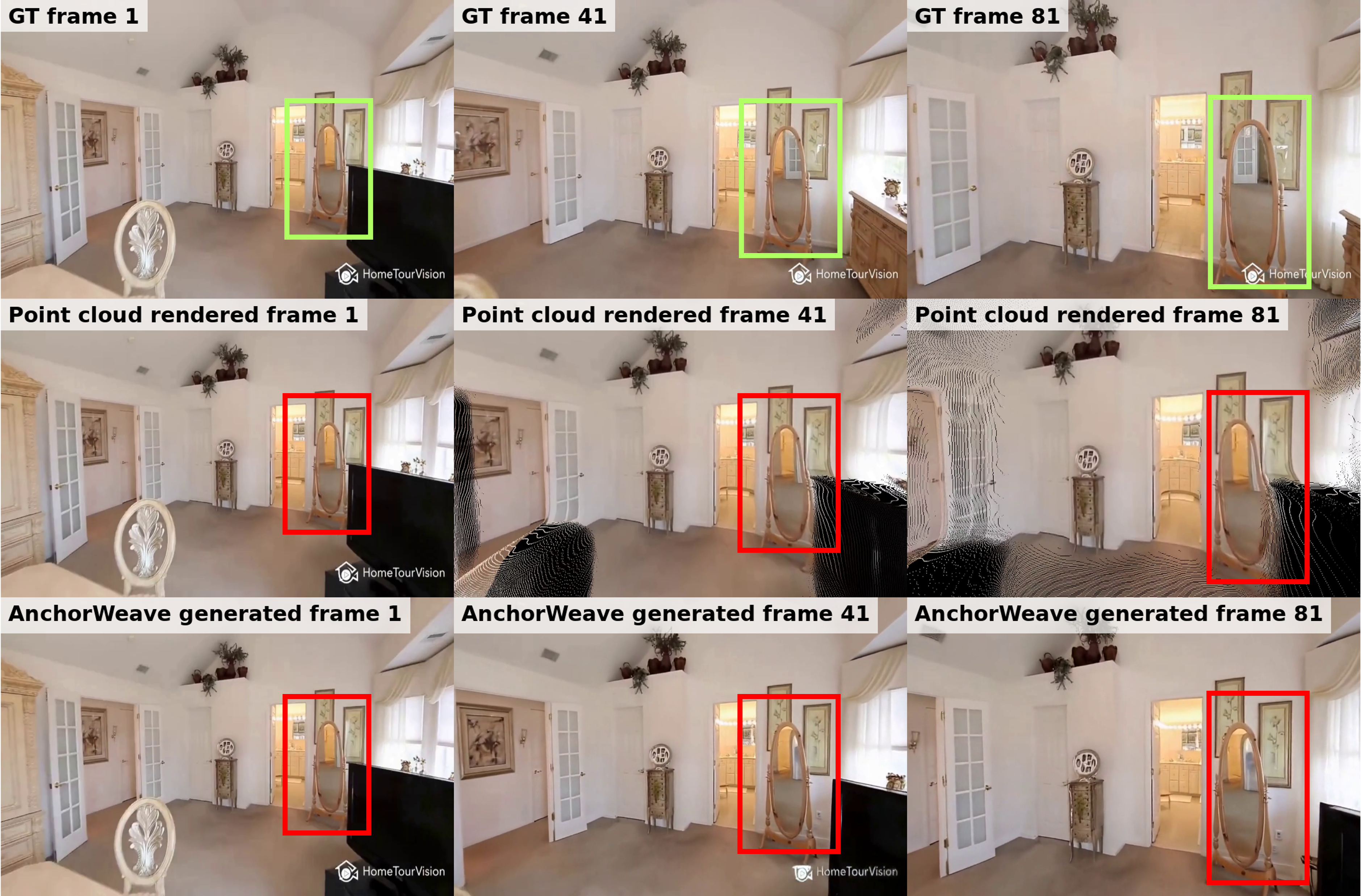}
    \caption{Reflective surfaces are a representative failure case: static geometry stores a fixed reflection, while the correct reflection is view-dependent.}
    \label{fig:mirror_failure}
\end{figure}

The analyses above also identify cases where static spatial memory remains challenging. Extremely long trajectories can accumulate synthesis errors and pose drift, as shown in \Cref{fig:long_stress}. Reflective surfaces are another representative failure mode: a point cloud stores the appearance observed from a particular viewpoint, but the correct reflection changes with camera motion. As shown in \Cref{fig:mirror_failure}, this can make the rendered memory inconsistent with the target view. Out-of-distribution scenes may also degrade the monocular 3D estimator, producing weaker anchors and less reliable conditioning.

\section{Limitations}
AnchorWeave improves spatial conditioning robustness but remains dependent on the capabilities of the underlying video diffusion backbone and 3D reconstruction models. 
While local geometric memories mitigate global fusion artifacts, residual errors in appearance synthesis or geometric estimation may still be written into memory and revisited in later segments. 
In challenging scenarios with large viewpoint extrapolation or complex dynamics, such imperfections may gradually accumulate over long horizons.

From an efficiency perspective, the dominant computational cost arises from the iterative denoising process of the video diffusion model. 
Thus, inference speed remains bounded by the backbone generator. 
Future work may explore distillation or consistency-based acceleration to further improve generation efficiency.

\end{document}